\documentclass[dvipsnames]{article} %
\usepackage{colm2024_conference}

\usepackage{booktabs}
\usepackage{enumitem}
\usepackage{wrapfig}
\usepackage{algorithm}
\usepackage{algpseudocode}
\usepackage{graphicx}
\usepackage{subfigure}
\usepackage[misc]{ifsym}
\usepackage{pdfpages}
\usepackage{amsmath}

\usepackage{fancyvrb}

\usepackage{listings}

\usepackage{microtype}
\usepackage{amsmath}
\usepackage{amssymb}
\usepackage{stmaryrd}
\usepackage{colortbl}
\usepackage[utf8]{inputenc}
\usepackage[T1]{fontenc}
\definecolor{lightgray}{rgb}{0.9,0.9,0.9}
\usepackage{caption}
\usepackage{subcaption}
\usepackage{setspace}
\usepackage{url}
\usepackage{multirow}
\usepackage{tabularx}
\usepackage{blindtext}
\usepackage{pgfplots}
\pgfplotsset{compat=1.18} 
\usepackage{tikz}
\usetikzlibrary{er,positioning,bayesnet}
\usepackage{makecell}
\usepackage{tipa}
\usepackage{siunitx}
\usepackage{nicefrac}
\usepackage{listings}
\usepackage[raster,skins, most]{tcolorbox} %
\usepackage{xltabular}
\usepackage{adjustbox}
\usepackage{xurl}
\usepackage{rotating}
\usepackage[normalem]{ulem}
\usepackage{booktabs}
\usepackage{fontawesome}

\useunder{\uline}{\ul}{}

\usepackage{amsmath,amsfonts,bm}

\def\eqref#1{equation~\ref{#1}}

\def\1{\bm{1}}

\DeclareMathAlphabet{\mathsfit}{\encodingdefault}{\sfdefault}{m}{sl}
\SetMathAlphabet{\mathsfit}{bold}{\encodingdefault}{\sfdefault}{bx}{n}

\newcommand*\justify{%
  \fontdimen2\font=0.4em
  \fontdimen3\font=0.2em
  \fontdimen4\font=0.1em
  \fontdimen7\font=0.1em
  \hyphenchar\font=`\-
}

\renewcommand{\texttt}[1]{%
  \begingroup
  \ttfamily
  \begingroup\lccode`~=`/\lowercase{\endgroup\def~}{/\discretionary{}{}{}}%
  \begingroup\lccode`~=`[\lowercase{\endgroup\def~}{[\discretionary{}{}{}}%
  \begingroup\lccode`~=`.\lowercase{\endgroup\def~}{.\discretionary{}{}{}}%
  \catcode`/=\active\catcode`[=\active\catcode`.=\active
  \justify\scantokens{#1\noexpand}%
  \endgroup
}

\usepackage{makecell}
\usetikzlibrary{tikzmark}
\makeatletter
\newcommand*\myfontsize{%
  \@setfontsize\myfontsize{7}{8}%
}
\makeatother

\definecolor{uclablue}{RGB}{159, 195, 224}

\definecolor{uclagold}{RGB}{255, 240, 180}

\definecolor{aliceblue}{RGB}{255, 238, 241}

\definecolor{cadmiumgreen}{rgb}{0.0, 0.42, 0.24}

\definecolor{myred}{rgb}{0.7, 0.3, 0.0}
\definecolor{myblue}{rgb}{0.2, 0.3, 0.6}
\definecolor{babygreen}{rgb}{0.85, 0.97, 0.85}

\definecolor{purple1}{RGB}{126, 107, 196}
\definecolor{purple2}{RGB}{199, 158, 207}
\definecolor{purple3}{RGB}{214, 200, 255}
\definecolor{purple4}{RGB}{254, 240, 255}

\definecolor{deepblue}{RGB}{48, 58, 82}

\hypersetup{ breaklinks, colorlinks=true, allcolors=uclablue_old }

\definecolor{deepPurple}{HTML}{330066}

\definecolor{uclablue_old}{rgb}{0.329, 0.318, 0.961}
\hypersetup{
    breaklinks,
    citecolor=uclablue_old,
    colorlinks=true,
}

\newtcolorbox{mybox}[2][]
  {colback = black!5!white, colframe = black!75!black, fonttitle = \bfseries,
    colbacktitle = black!100!black, enhanced, before upper={\fontsize{8}{11}\obeyspaces\obeylines\selectfont}, fontupper=\selectfont,
    attach boxed title to top left={yshift=-2.2mm,xshift=4mm},
    title=#2,#1}

\title{%
  \begin{tabular}[t]{l}
  \parbox[t]{\textwidth}{ 
    From Context to EDUs: Faithful and Structured Context Compression via Elementary Discourse Unit Decomposition
  }
  \end{tabular}
}

\author{
\textbf{Yiqing Zhou\textsuperscript{*}$^{\heartsuit}$, Yu Lei\textsuperscript{*}$^{\heartsuit}$$^\diamondsuit$, Shuzheng Si\textsuperscript{*}$^{\spadesuit}$$^{\heartsuit}$, Qingyan Sun\textsuperscript{*}$^{\heartsuit}$, Wei Wang$^{\heartsuit}$$^{\bigstar}$\\ Yifei Wu$^{\heartsuit}$, Hao Wen$^{\heartsuit}$, Gang Chen$^{\heartsuit}$, Fanchao Qi$^{\heartsuit}$$^{\spadesuit}$$^{(\textrm{\Letter})}$, Maosong Sun$^{\spadesuit}$}
\\[0.5em]
{\fontsize{10pt}{11pt}\selectfont
$^{\heartsuit}$ DeepLang AI \\
$^{\spadesuit}$ Department of Computer Science and Technology, Tsinghua University \\ $^\diamondsuit$ Beijing University of Posts and Telecommunications \\ $^{\bigstar}$ Beijing Jiaotong University 
}
}

\begin{document}
\vspace*{-0.8cm}

\maketitle

\begingroup
  \renewcommand\thefootnote{*}  
  \footnotetext{~Equal contribution.}
\endgroup

\begingroup
  \renewcommand\thefootnote{\Letter}  
  \footnotetext{~Corresponding author. Emails: fanchao.qi@deeplang.ai}
\endgroup

\vspace{-0.1in}
\begin{abstract}
Managing extensive context remains a critical bottleneck for Large Language Models (LLMs), particularly in applications like long-document question answering and autonomous agents where lengthy inputs incur high computational costs and introduce noise. Existing compression techniques often disrupt local coherence through discrete token removal or rely on implicit latent encoding that suffers from positional bias and incompatibility with closed-source APIs. To address these limitations, we introduce the EDU-based Context Compressor, a novel explicit compression framework designed to preserve both global structure and fine-grained details. Our approach reformulates context compression as a structure-then-select process. First, our LingoEDU transforms linear text into a structural relation tree of Elementary Discourse Units (EDUs) which are anchored strictly to source indices to eliminate hallucination. Second, a lightweight ranking module selects query-relevant sub-trees for linearization. To rigorously evaluate structural understanding, we release StructBench, a manually annotated dataset of 248 diverse documents. Empirical results demonstrate that our method achieves state-of-the-art structural prediction accuracy and significantly outperforms frontier LLMs while reducing costs. Furthermore, our structure-aware compression substantially enhances performance across downstream tasks ranging from long-context tasks to complex Deep Search scenarios.

\end{abstract}

\begin{figure}[H]
    \centering
    \includegraphics[width=0.95\textwidth]{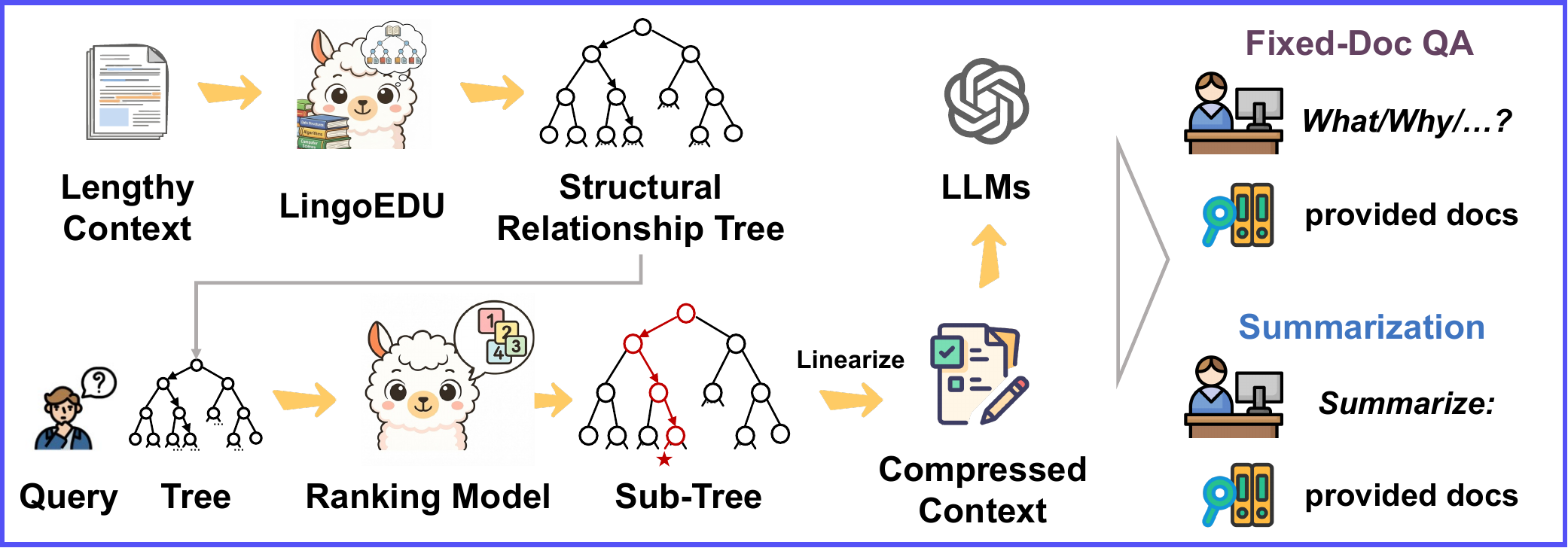}
    \caption{Overview of the EDU-based Context Compressor framework.}
    \label{fig:intro}
\end{figure}


\section{Introduction}

Large Language Models (LLMs) have achieved remarkable progress in recent years \citep{gpt4, llama3, deepseekai2025deepseekv3technicalreport, qwen3}, demonstrating capabilities that approach or surpass human performance on diverse tasks \citep{chen2021dialogsum, si-etal-2022-scl, si2023spokenwoz, yi2024survey, luo2025large, an-etal-2025-ultraif}. 
These advances have empowered sophisticated applications such as long-context question answering (QA) \citep{si-etal-2025-gateau}, multi-document summarization \citep{kryściński2022booksumcollectiondatasetslongform}, and autonomous agents capable of performing deep research and complex reasoning \citep{lei2025rhinoinsight, si2025goalplanjustwish}.
However, these applications typically require maintaining an extensive memory for context information, e.g., the used documents and agent-environment interactions. 
As the capabilities of LLMs continue to grow and their application scenarios expand, effectively managing such contextual memory becomes a critical bottleneck for long-context tasks. 
The accumulated and lengthy context not only incurs prohibitive computational costs but also introduces significant noisy information, which can overwhelm the model and degrade the final performance of LLMs \citep{hua2025contextengineering20context, wang-etal-2025-document}.

To better organize the lengthy context, context compression has emerged as a crucial technique to reduce input token length while preserving maximal semantic integrity.
\textbf{Explicit compression methods} \citep{jiang2023llmlingua, xu2024concise} attempt to reduce the length of context by removing tokens or sentences deemed less important, e.g., using an abstractive summarization model as a compressor to obtain the shorter context and maintain global meaning \citep{xu2023recompimprovingretrievalaugmentedlms}. 
However, these methods often operate on discrete tokens or rigid sentence boundaries, disrupting the local coherence of the text. 
Meanwhile, they typically focus on preserving the most important global information, while overlooking the original article’s structural information and fine-grained details.
Conversely, \textbf{implicit compression methods} \citep{mu2024learning, gecontext, liu2025contextcascadecompressionexploring} try to encode lengthy text into latent vectors to achieve higher compression ratios.
However, recent studies \citep{li2025admtreecompressinglengthycontext} show that implicit compression methods tend to have positional bias. 
This means they often ignore information from the beginning or middle of the context, focusing instead on the most noticeable content and overlooking less prominent details.
Also, these implicit methods \citep{cheng2025glyphscalingcontextwindows, wei2025deepseekocrcontextsopticalcompression} tend to lack flexibility, as they often require specially designed post-training processes or the use of latent vectors as new inputs. 
This limits the applicability of such techniques to advanced API-based models, e.g., GPT-4.1 \citep{gpt-4.1}.

In this work, we posit that an ideal compression strategy for long-context scenarios should be \textit{explicit} to ensure flexibility while focusing on the structural information of the original context, thereby maintaining both global foresight and fine-grained details.
Our design goal is to first transform a linear context sequence into a structural relation tree, where each node is strictly anchored to the source via coordinate pointers.
Subsequently, we select the sub-tree relevant to the input query, then linearize it as the compressed context.
In this way, our explicit compression method can be guided to retain not only the most salient global information, but also the structural relationships, fine-grained details, and coherent sentences, which are essential for faithful downstream reasoning.
Therefore, we introduce the EDU-based Context Compressor, which consists of the LingoEDU and a lightweight ranking module.
Specifically, the LingoEDU is inspired by rhetorical structure theory and built upon elementary discourse units (EDUs) \citep{mann1988rhetorical}.
Unlike fixed tokens or sentences, EDUs are the minimal variable-length units that coherently convey one piece of information.
LingoEDU aims to transform unstructured context sequences into structural relationship trees, where nodes represent EDUs and edges represent the existence of a discourse linkage and dependency relation between two EDUs.
To efficiently obtain the structural relationship trees, we train the LingoEDU via a novel human-in-the-loop pipeline to obtain the training data followed by a supervised fine-tuning (SFT) stage.
After obtaining the structural relation tree through the EDU-based decomposition module, we subsequently employ a ranking module to get the most useful sub-trees to achieve context compression.
By taking the task instruction and the structural relation tree as input, we use a ranking model to identify task-relevant sub-trees and then linearize them into a compressed context.
For instance, given a multi-doc QA task, the EDU-based decomposition module first maps the lengthy context into a structural relationship tree where nodes represent semantic EDUs and edges encode their original positional references. 
Then, the ranking model returns the nodes that are highly relevant to the user query, filtering out noisy content at the structural level. 
This step effectively prunes the full document tree into a query-specific subtree, retaining only the most relevant EDUs. 
Finally, this refined sub-tree is linearized and fed into the target LLM alongside the original query to generate the final response.
This design allows our method to preserve global structure while capturing fine-grained details, and crucially, since it operates without latent representations, it ensures seamless compatibility with API-based models.

During the experiments, we aim to answer two pivotal questions: \textbf{(1) Can the state-of-the-art LLMs effectively compress long contexts while preserving original structural information?} and \textbf{(2) Does such structure-aware compression tangibly reduce the hallucinations for downstream tasks?} 
To answer these questions, we first propose a manually annotated benchmark comprising 248 documents across diverse formats to evaluate the abilities of LLMs to understand and describe structural information of the context.
We find that even state-of-the-art models such as o3 \citep{o3} still fail to fully understand the structural relationships within a given context, and prompt-based methods alone for these models struggle to effectively compress the context while preserving the original structural relations.
Conversely, our well-trained EDU-based Context Compressor can effectively identify and preserve key structural elements while substantially reducing the context length.
Meanwhile, we find that using the structured and compressed context from EDU-based Context Compressor not only reduces the input length for the model, but also filters out irrelevant and noisy tokens, thereby improving the model’s performance and reducing hallucinations on various tasks such as multi-document QA \citep{bai2023longbench}.


Our contributions are summarized as follows:
\begin{itemize}
    \item \textbf{Novel Context Compression Framework:}  We introduce the EDU-based Context Compressor that leverages  document structure to create concise, informative sub-trees tailored to specific queries. 
    This approach preserves both global structure and local details for context compression. Also, it remains fully compatible with closed-source models like GPT-4.1.
    
    \item \textbf{New Benchmark:} We release a manually annotated benchmark comprising 248 documents across diverse formats to enable precise evaluation of the abilities of LLMs to understand and describe structural information of the context.
    
    \item \textbf{SOTA Performance \& Efficiency:} Empirical results show that our method significantly outperforms advanced LLMs (e.g., o3) in understanding structural relations within the provided context and surpasses commercial APIs like Firecrawl, notably with a much lower cost.
    
    \item \textbf{Reducing Hallucinations for Long-context Tasks:} We demonstrate that our proposed EDU-based Context Compressor can improve the final performance and reduce hallucinations across diverse long-context tasks, e.g., multi-document QA, summarization, and search agent scenarios.
    For example, our method outperforms standard baselines (e.g., +14.94\% on HotpotQA within LongBench) by preserving precise evidence chains. In Deep Search tasks, it significantly enhances the performance, boosting DeepSeek-R1 by over \textbf{51.11\%} relatively on the HLE benchmark.
\end{itemize}

\section{Methodology}

We propose the \textbf{EDU-based Context Compressor}, a novel framework designed to achieve faithful and structural-aware context compression. As argued in the Introduction, implicit processing often suffers from positional bias and lack of transparency. Therefore, our approach adheres to an \textbf{explicit compression paradigm}: it transforms the linear context sequence into a \textbf{Structural Relation Tree}, where semantic units are strictly anchored to the source text via coordinate pointers.
As illustrated in Figure~\ref{fig:intro}, the framework operates as a plug-and-play module compatible with any LLM (including API-based models). It consists of two cascaded components: the \textit{LingoEDU}, which parses the document into discourse-connected units, and a \textit{Ranking Module}, which identifies and linearizes the most relevant sub-trees to reconstruct a high-density context.

\subsection{Overall Framework}
\label{sec:overall}

Given a long input document $\mathcal{D}$ (or a set of documents) and a user query $q$, our goal is to overcome the limitations of fixed-size chunking and latent encoding. We reformulate the context compression task as a \textbf{``Structure-then-Select''} process:

\begin{enumerate}
    \item \textbf{Phase I: Structural Decomposition.} The Decomposer transforms the unstructured linear text $\mathcal{D}$ into a \textbf{Structural Relation Tree} $\mathcal{T} = (\mathcal{V}, \mathcal{E})$.
    \begin{itemize}
        \item \textbf{Nodes ($\mathcal{V}$):} represent \textbf{Elementary Discourse Units (EDUs)}, the minimal variable-length units capable of conveying coherent semantics. Crucially, strictly preserving the original text indices ensures hallucination-free hallucination.
        \item \textbf{Edges ($\mathcal{E}$):} represent the \textit{discourse linkages} and dependency relations between EDUs (e.g., elaboration, contrast), capturing the logical flow often lost in standard retrieval.
    \end{itemize}
    
    \item \textbf{Phase II: Sub-tree Retrieval and Linearization.} A lightweight ranking module evaluates the relevance between the query $q$ and the structural nodes in $\mathcal{T}$. Instead of retrieving isolated sentences, it identifies the optimal task-relevant \textbf{sub-trees} $\mathcal{S} \subset \mathcal{T}$. Finally, these selected sub-trees are \textbf{linearized} back into a coherent text sequence $\mathcal{D}'$. This results in a compressed context where $|\mathcal{D}'| \ll |\mathcal{D}|$, retaining both global structural integrity and fine-grained details essential for downstream reasoning.
\end{enumerate}

\subsection{LingoEDU}
\label{sec:decomposer}
\begin{figure*}[h]
    \centering
    \includegraphics[width=0.98\textwidth]{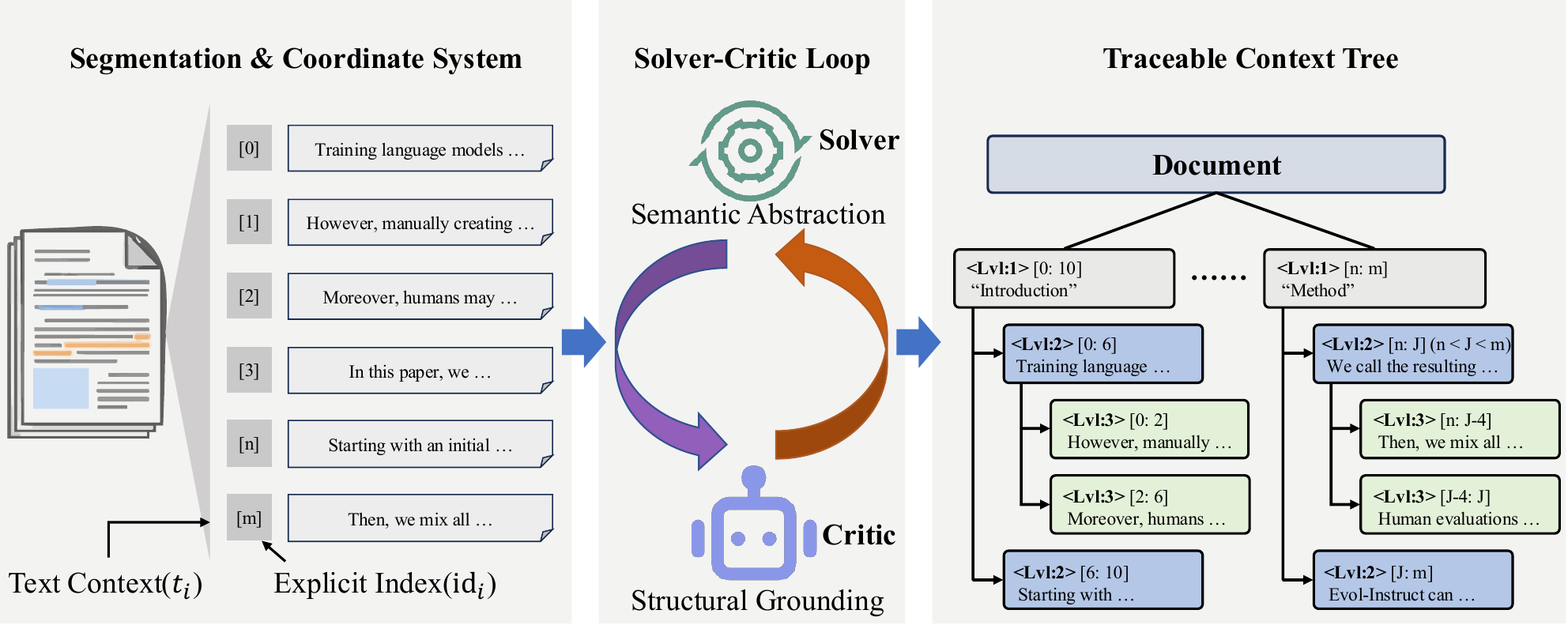}
    \caption{\textbf{Overview of the LingoEDU.} 
    (a) \textbf{Coordinate System Construction}: The continuous input text is segmented into EDUs, creating an addressable sequence where each unit carries a unique coordinate ID. 
    (b) \textbf{Traceable Generation}: Unlike standard summarization, the model outputs \textit{Augmented Markdown}. It performs compression by generating closed index intervals (e.g., $[12\text{--}15]$) rather than regenerating body text, effectively indexing the content. 
    (c) \textbf{Tree Realization}: The output is parsed into a hierarchical semantic tree $\mathcal{T}$. The explicit ID spans serve as unhallucinated anchors, ensuring the abstractive nodes remain strictly faithful to the source context.}
    \label{fig:method}
\end{figure*}

We frame the task of Document Structure Analysis as a \textbf{traceable context compression} problem. As illustrated in Figure~\ref{fig:method}, our framework operates by transforming a linear discourse sequence into a condensed hierarchical tree, where every node is strictly anchored to the source via coordinate pointers.

\subsubsection{Problem Formulation: Coordinate-Based Discourse Representation}
To enable this coordinate-based operation (Figure~\ref{fig:method}(a)), we first segment the input document $\mathcal{D}$ into a sequence of atomic building blocks termed \textbf{Elementary Discourse Units (EDUs)}. Formally, the document is represented as a sequence $\mathcal{U} = \{e_1, e_2, \dots, e_N\}$, where each unit $e_i$ constitutes a triplet:
\begin{equation}
    e_i = \left( t_i,  \mathtt{pos}_i, \mathtt{id}_i \right)
\end{equation}
Here, $t_i$ represents the textual content (typically a coherent clause or sentence), $\mathtt{pos}_i$ denotes its physical grounding (character offsets), and crucially, $\mathtt{id}_i$ acts as a unique sequential index. 
This index establishes a \textbf{Coordinate System}, allowing the system to reference content by pointers.
The core objective is to learn a decomposition function $f: \mathcal{E} \rightarrow \mathcal{T}$ that maps the linear EDU sequence to a hierarchical tree $\mathcal{T}$ (Figure~\ref{fig:method}(c)). Each node $n_j \in \mathcal{T}$ acts as a compressed semantic capsule defined as:
\begin{equation}
    n_j = \left( h_j, l_j, \sigma_j \right), \quad \text{where } \sigma_j = [\mathtt{id}_{\text{start}}, \mathtt{id}_{\text{end}}]
\end{equation}
In this tuple, $h_j$ is the semantic abstract (e.g., a section title or summary), $l_j$ denotes the hierarchical depth, and $\sigma_j$ represents the \textbf{EDU Span}—a closed interval explicitly pointing to the source range in $\mathcal{E}$. By enforcing the constraint $\sigma_j \subseteq [1, N]$, we ensure \textbf{referential integrity}: the generated structure is a lossless index purely derived from the input context, effectively eliminating generative hallucinations.

\subsubsection{Training Strategy and Data Synthesis}
Since high-quality, fine-grained hierarchical annotations for long contexts are scarce, we introduce a scalable automated pipeline to synthesize training data. This pipeline leverages a strong LLM to distill the logic of summarization-and-indexing.

\paragraph{Bi-Level Task Decomposition.}
Recognizing that faithfulness requires different cognitive capabilities for different structural types, we decouple the data generation into two distinct sub-tasks to prevent ``instruction conflict'' (conflating visual layout with semantic reasoning):
\begin{enumerate}
    \item \textbf{Explicit Layout Extraction:} The model extracts objective structural cues (e.g., Markdown headers, HTML tags) to form the document skeleton. This task enables high certainty with low ambiguity.
    \item \textbf{Deep Semantic Segmentation:} For large text blocks lacking explicit formatting, the model focuses purely on semantic shifts to delineate finer-grained functional sections. This requires deep reasoning to resolve high ambiguity.
\end{enumerate}

\paragraph{The Solver-Critic Refinement Loop.}
To ensure the synthesized labels are high-quality, we implement an \textit{Iterative Refinement Mechanism}:
\begin{itemize}
    \item \textbf{The Solver:} Proposes a hierarchical decomposition, attempting to abstract detailed content into high-level semantic nodes.
    \item \textbf{The Critic:} Audits the proposal specifically checking whether the generated abstract (Title) accurately reflects the assigned span $\sigma_j$ without hallucination or semantic drift.
\end{itemize}
This adversarial collaboration ensures that despite the high compression rate, the structural integrity of the training data remains intact.

\subsubsection{Traceable Generation}
We train an open-source LLM on this synthesized corpus. To further guarantee robustness during inference, we employ an \textbf{Augmented Markdown Schema} (Figure~\ref{fig:method}(b)). The model is trained to generate nodes in the following format:
\begin{equation}
    \text{Output} = \underbrace{\text{\#\#}}_{\text{Level}} \ \underbrace{[\text{id}_{\text{start}}\text{--}\text{id}_{\text{end}}]}_{\text{Traceable Anchor}} \ \underbrace{\text{Concept Title}}_{\text{Semantic Abstract}}
\end{equation}
This design achieves dual goals: (1) \textbf{Token Efficiency}, as long text blocks spanning multiple EDUs are compressed into minimal tokens; and (2) \textbf{Hallucination Elimination}, as the strict span format forces the model to rely on the coordinate system rather than free-form generation. Constraints are further enforced during decoding to ensure only valid numerical indices from $\mathcal{E}$ are generated.

\subsection{Ranking Module}
\label{sec:rerank}

Once the document is decomposed into the structural tree $\mathcal{T}$, the challenge shifts to identifying which nodes contribute most to answering the user query $q$. We introduce a \textit{Budget-Aware Semantic Filter} that leverages the Decomposer's output to perform precise context selection.

\subsubsection{Why Node-Level Ranking?}
Instead of retrieving at the sentence level—which often results in context fragmentation—or the document level—which introduces excessive noise—we perform retrieval at the \textit{Node Level}. Our nodes $n_j$ encapsulate semantic completeness via their spans, allowing the model to judge the relevance of larger logical blocks without processing the full text.

\subsubsection{Plug-and-Play Relevance Scoring}
We define a relevance scoring function $\phi(q, n_j)$ to quantify the pertinence of each node. While our framework allows $\phi$ to be instantiated by state-of-the-art LLMs (e.g., GPT-4) via prompt-based scoring, such approaches are computationally prohibitive for scanning dense tree structures. 
To balance performance with efficiency, we employ a {lightweight ranking model} as a cost-effective surrogate. Specifically, we utilize an open-source model\footnote{\url{https://huggingface.co/Qwen/Qwen3-Reranker-0.6B}} to compute the relevance score $s_j$:
\begin{equation}
    s_j = \phi_{\theta}\left(q, \; h_j \oplus t_{\text{rep}}\right)
\end{equation}
where $h_j$ is the generated abstract (title) and $t_{\text{rep}}$ is the representative text snippet from the span $\sigma_j$. By using a compact model (e.g., 0.6B parameters) rather than large-scale LLM APIs, we achieve high-throughput filtering with negligible latency.

\subsubsection{Budget-Aware Greedy Selection}
To address the limitations of fixed Top-$K$ retrieval (which neglects token consumption), we employ a dynamic selection strategy bounded by a context budget $B_{\text{max}}$.

We sort all nodes in $\mathcal{T}$ by their score $s_j$ in descending order and select nodes into a candidate set $\mathcal{C}$:
\begin{equation}
    \mathcal{C} = \left\{ n_j \mid \sum_{n \in \mathcal{C}} \text{Len}(\text{Retrieve}(\sigma_n)) \leq B_{\text{max}} \right\}
\end{equation}
where $\text{Retrieve}(\sigma_n)$ fetches the original text EDUs corresponding to the span $[\mathtt{id}_{\text{start}}, \mathtt{id}_{\text{end}}]$. This greedy strategy aligns the retrieved context density with the LLM's optimal window size.

\subsubsection{Linearization}
A critical failure mode in standard RAG is the loss of discourse coherence when disjoint chunks are concatenated. Thanks to the explicit coordinates $\mathtt{id}$ provided by our Decomposer, we apply a \textbf{Re-ordering Protocol}. The selected spans in $\mathcal{C}$ are sorted by their original start indices $\mathtt{id}_{\text{start}}$ before concatenation. This restoration of logical order enables the downstream LLM to perform effective reasoning across discontinuous but structurally organized segments.

\section{Experiments}
\label{sec:experiments}

\subsection{Evaluation of Structural Integrity and Compression}
\label{sec:exp_structure}

In this section, we address the first research question: \textit{Can state-of-the-art LLMs effectively compress long contexts while preserving original structural information?} We compare our proposed method against frontier LLMs and commercial parsing APIs on a newly constructed benchmark.

\subsubsection{Experiment Settings}
\paragraph{Benchmark Construction.}
The absence of public benchmarks for fine-grained document structure analysis motivated us to construct a specific dataset named \textbf{StructBench}. We compiled a test set of 248 documents, covering diverse formats (Web pages, PDFs), languages (Chinese, English), and genres (e.g., government files, institutional reports, academic papers, and technical tutorials). 
The dataset spans 10 distinct genres, primarily focusing on complex structures such as academic papers, government files, business reports, and technical tutorials (full distribution in Appendix~\ref{sec:dataset_details}).
Document lengths vary significantly, ranging from 300 to 50,000 words.
To ensure high-quality ground truth, documents were parsed, sentence-segmented, and manually annotated for discourse structure by human experts. To enable fair comparison with baselines that may struggle with leaf-level details, we extracted the \textit{structural backbone} (top-level hierarchy) from the annotations to serve as the labels.
\footnote{The dataset is available at \url{https://huggingface.co/datasets/deeplang-ai/StructBench}}

\begin{table*}[h]
    \centering
    \resizebox{0.83\textwidth}{!}{
    \begin{tabular}{l|c|cc|cc}
    \toprule
    \textbf{Method} & \textbf{Type} & \textbf{TED (Structure)}~$\bm{\downarrow}$ & \textbf{DLA (Accuracy)}~$\bm{\uparrow}$ & \textbf{Cost (\$)}~$\bm{\downarrow}$ & \textbf{Latency (s)}~$\bm{\downarrow}$ \\
    \midrule
    GPT-4o & \multirow{12}*{General LLM$^\ast$} & ~~6.22 & 29.03\% & 5.21 & - \\
    GPT-4.1 & ~ & ~~6.35 & 37.90\% & 4.17 & - \\
    OpenAI o3 & ~ & ~~5.51 & 28.63\% & 4.17 & - \\
    OpenAI o4-mini & ~ & ~~5.87 & 32.66\% & 2.28 & - \\
    Claude-3.7-Sonnet & ~ & ~~6.65 & 35.08\% & 7.09 & - \\
    Claude-4-Sonnet & ~ & ~~\underline{5.08} & \underline{43.15\%} & 7.09 & - \\
    Gemini-2.5-Flash & ~ & ~~5.82 & 27.82\% & 0.99 & - \\
    Gemini-2.5-Pro & ~ & ~~5.61 & 32.66\% & 4.02 & - \\
    DeepSeek-V3 & ~ & ~~6.32 & 33.47\% & 0.30 & - \\
    DeepSeek-R1 & ~ & ~~6.26 & 30.65\% & 1.14 & - \\
    Qwen3-32B & ~ & ~~9.49 & 24.90\% & 0.26 & 10.17$^\dagger$ \\
    Qwen3-235B & ~ & ~~9.93 & 17.89\% & 0.11 & - \\
    \midrule
    Jina-Reader & \multirow{2}*{Parser API} & 17.04 & - & \textbf{0.10} & - \\
    Firecrawl & ~ & 16.81 & - & \underline{0.17} & - \\
    \midrule
    \rowcolor{gray!10} \textbf{Our Method (LingoEDU)} & \textbf{Specialized} & ~~\textbf{4.77} & \textbf{49.60\%} & \underline{0.17} & ~~\textbf{1.20}$^\dagger$ \\
    \bottomrule
    \end{tabular}
    }
    \caption{Performance comparison on \textbf{StructBench}. $\ast$ indicates the model is accessed via API. $\dagger$ denotes local deployment for latency testing using equivalent hardware. Costs are calculated for the entire test set (approx. 248 documents). Best results are \textbf{bolded}, and second-best are \underline{underlined}.}
    \label{tab:main_table}
\end{table*}

\paragraph{Evaluation Metrics.}
We employ two complementary metrics to evaluate structural fidelity. \textbf{Tree Edit Distance (TED)}~\citep{ted_zhang1989simple} acts as a micro-level metric to measure structural dissimilarity by computing the minimum number of edit operations (insertion, deletion, substitution) required to transform the predicted tree into the ground truth, where a lower TED indicates more precise structural alignment. Complementarily, \textbf{Document Level Accuracy (DLA)} serves as a macro-level metric defined as $\mathrm{DLA} = \frac{\lvert D_{\text{match}} \rvert}{\lvert D_{\text{all}} \rvert}$, in which $D_{\text{match}}$ represents the count of documents where the decomposed structural backbone perfectly matches the ground truth. This rigorous metric requires zero structural errors.

\paragraph{Baselines.}
We compare our \textbf{LingoEDU} (built on Qwen3-4B~\citep{qwen3technicalreport}) against two categories of strong baselines. For all LLMs, we designed specific prompts to instruct them to output hierarchical JSON/Markdown structures:
(1)\textbf{Frontier LLMs:} We evaluated SOTA models including GPT-4o/4.1~\citep{gpt-4o}, OpenAI o3/o4-mini~\citep{o3}, Claude 3.7 Sonnet/4 Sonnet~\citep{Claude4}, DeepSeek-V3/R1~\citep{deepseekai2025deepseekv3technicalreport}, and Qwen3-235B~\citep{qwen3}. All LLM results are averaged over three distinct runs.
(2)\textbf{Commercial Parsing APIs:} We selected Jina Reader and Firecrawl, which are widely used for web-to-markdown conversion. We deployed test documents on a static server to allow URL-based access.

\paragraph{Implementation Details.}
Our method utilizes Qwen3-4B as the backbone. The training involved a two-stage process: (1) \textbf{Continued Pre-training} on $\sim$100k synthetic samples to learn layout patterns, followed by (2) \textbf{Supervised Fine-Tuning (SFT)} on thousands of meticulously manually annotated documents to align with human intent.
All experiments were conducted on a Linux operating system running on a high-performance server equipped with an Intel Xeon 2.3GHz CPU, 1960GB of memory, and 8 NVIDIA A100 GPUs, each with 80 GB of VRAM.

\subsubsection{Experiment Results}

\paragraph{Analysis of Structural Integrity.}
Table~\ref{tab:main_table} highlights the superiority of our explicit training paradigm. While top-tier commercial LLMs like Claude-4-Sonnet and OpenAI o3 achieve competitive structural scores (TED $\sim$5.1--5.5), other robust models such as DeepSeek-R1 and Qwen3 still struggle, plateauing at a TED of 6.2--9.9. Qualitative analysis reveals that general models often \textit{hallucinate} non-existent sub-sections or flatten deep hierarchies to save generation tokens. Similarly, commercial parsing APIs lack semantic depth; Jina and Firecrawl exhibit high TED scores ($>16$) as they rely on shallow HTML tags and fail to capture implicit discourse structures found in complex PDFs. In contrast, our LingoEDU Decomposer demonstrates specialized efficiency by achieving a remarkable TED of \textbf{4.77} and a DLA of \textbf{49.60\%}, significantly outperforming the strongest baseline, Claude-4-Sonnet (+\textbf{6.45\%} absolute DLA). This confirms that structural understanding requires dedicated supervision beyond what emergent prompting or reasoning models can provide.

\paragraph{Efficiency and Cost Analysis.}
In real-world long-context applications, overhead is critical. As shown in Table~\ref{tab:main_table}, our method offers an optimal trade-off. It matches the cost of the cheapest parsers (\$0.0007/doc) while delivering a latency of just 1.20 seconds per document—nearly \textbf{10$\times$ faster} than a locally deployed Qwen3-32B. This efficiency stems from our architecture's design, which outputs compact coordinate indices instead of generating verbose text.

\subsubsection{Ablation Studies}

Table~\ref{tab:ablation} validates the effectiveness of our design choices. 
First, the significant performance gap between ``Indices Only'' and our method highlights that explicit text generation acts as a crucial semantic anchor for structural prediction. 
Second, the model exhibits remarkable data efficiency; even when scaled down to just \textbf{20\%} of the training data, it achieves a TED of 4.87 and retains over \textbf{91\%} of the full model's accuracy.

\begin{table}[h]
    \centering
    \resizebox{0.48\textwidth}{!}{
    \begin{tabular}{l|c|c|c}
    \toprule
    \textbf{Ablation Setting} & \textbf{Variant} & \textbf{TED} & \textbf{DLA (\%)} \\
    \midrule
    \multirow{2}{*}{\textit{Output Formulation}} & Indices Only & 8.16  & 33.06 \\
    & \textbf{Indices + Text (Ours)} & \textbf{4.77}  & \textbf{49.60} \\
    \midrule
    \midrule
    \multirow{3}{*}{\textit{Data Scaling}} & 20\% Data  & 4.87 & 45.16 \\
    & 50\% Data & 4.85 & 48.79 \\
    & \textbf{100\% Data} & \textbf{4.77} & \textbf{49.60} \\
    \bottomrule
    \end{tabular}
    }
     \caption{{Ablation Studies.} We analyze the impact of formulation and training data scale.}
     \label{tab:ablation}
\end{table}

Finally, we investigate the impact of model scale on structural parsing capability, as shown in Table~\ref{tab:ablation_scale}. Scaling the backbone from 1.7B to 4B yields clear improvements, reducing TED from 4.99 to 4.77. However, further scaling to 8B results in performance saturation: both TED and DLA regress compared to the 4B model. This suggests that the 4B parameter range strikes an optimal balance for this task, whereas larger models may suffer from overfitting to the rigid output format without proportionally larger datasets.

\begin{table}[t]
    \centering
    \small
    \setlength{\tabcolsep}{6mm}
    \begin{tabular}{lcc}
    \toprule
    \textbf{Model Size} & \textbf{TED} ($\downarrow$) & \textbf{DLA \%} ($\uparrow$) \\ 
    \midrule
    Qwen-1.7B & 4.99 & 48.39 \\
    Qwen-4B   & \textbf{4.77} & \textbf{49.60} \\
    Qwen-8B   & 4.89 & 49.19 \\
    \bottomrule
    \end{tabular}
    \caption{Ablation study on backbone model scaling. Our 4B model achieves the best balance between structural error (TED) and relation accuracy (DLA).}
    \label{tab:ablation_scale}
\end{table}

\begin{table*}[h]
\centering
\resizebox{\textwidth}{!}{%
\begin{tabular}{l|cccc|cccc|cccc}
\toprule
\multirow{2}{*}{\textbf{Model / Method}} & \multicolumn{4}{c|}{\textbf{Multi-Doc QA}} & \multicolumn{4}{c|}{\textbf{Summarization}} & \multicolumn{4}{c}{\textbf{Few-shot}} \\
 & HotpotQA & 2Wiki & Musique & DuReader & GovRep & QMSum & MultiN & VCSum & TREC & Trivia & SAMSum & LSHT \\ \midrule
\textbf{C3}&  0.07 & 0.09  & 0.08 & 2.08  & 18.20 & 7.35 & 18.03 & 0.39 & 1.00 & 6.42  & 8.29 & 6.50 \\
\textbf{Glyph} & 66.42 & 72.98 & - & - & 25.53 & 19.78 & - & - & \textbf{82.62} & 88.54 & - & - \\ \midrule
\multicolumn{13}{l}{\textit{\textbf{Gemini-2.5-Pro}}} \\
\quad Standard & 35.20 & 38.10 & 28.55 & 7.15 & 4.10 & 15.80 & 4.05 & 5.80 & 46.50 & 59.85 & 20.45 & 26.10 \\
\quad Self-Sum & 37.78 & 39.90 & 30.77 & 7.79 & {4.34} & {16.53} & 4.44 & 6.17 & 49.00 & 62.31 & 21.89 & 29.50 \\
\quad \textbf{Ours (LingoEDU)} & {40.46} & {40.91} & {31.22} & {8.12} & 4.25 & 16.17 & {4.85} & {6.36} & {57.50} & {63.25} & {23.80} & {35.48} \\
\quad \textcolor{blue}{$\Delta$ \textit{(vs. Standard)}} & \textit{+14.94\%} & \textit{+7.38\%} & \textit{+9.35\%} & \textit{+7.69\%} & \textit{+2.44\%} & \textit{+2.34\%} & \textit{+19.75\%} & \textit{+9.66\%} & \textit{\textbf{+23.66\%}} & \textit{+1.25\%} & \textit{+11.39\%} & \textit{+3.45\%} \\ \midrule
\multicolumn{13}{l}{\textit{\textbf{GPT-4.1}}} \\
\quad Standard & 65.83 & 72.98 & 51.90 & 21.80 & 29.97 & \underline{22.84} & \underline{20.85} & 12.50 & 77.00 & 90.07 & 39.20 & 48.60 \\
\quad Self-Sum & \underline{67.89} & \underline{74.39} & \underline{53.48} & \underline{23.51} & \underline{30.98} & 22.53 & {22.06} & \underline{13.71} & {79.00} & \underline{93.69} & \underline{40.79} & \underline{50.50} \\
\quad \textbf{Ours (LingoEDU)} & \textbf{70.11} & \textbf{74.68} & \textbf{54.86} & \textbf{25.34} & \textbf{31.56} & \textbf{23.30} & \textbf{23.50} & \textbf{14.62} & \underline{80.00} & \textbf{93.76} & \textbf{41.68} & \textbf{52.50} \\
\quad \textcolor{blue}{$\Delta$ \textit{(vs. Standard)}} & \textit{+6.50\%} & \textit{+2.33\%} & \textit{+5.70\%} & \textit{+16.24\%} & \textit{+2.94\%} & \textit{+0.61\%} & \textit{+5.80\%} & \textit{+8.96\%} & \textit{+3.90\%} & \textit{+4.10\%} & \textit{+6.33\%} & \textit{+8.02\%} \\ \bottomrule
\end{tabular}%
}
\caption{Results on \textbf{LongBench}. Datasets are grouped by task type (columns). $\Delta$ denotes the relative improvement of \textbf{Ours} over the Standard baseline for each specific backbone. \textbf{Bold} indicates best performance per backbone group; \underline{underlined} indicates second-best.}
\label{tab:main_results}
\end{table*}

\subsection{Main Results on Downstream Long-Context Tasks}
\label{sec:practical_utility}

In this section, we address the second research question: \textit{Does structure-aware compression tangibly enhance performance for downstream tasks?} We evaluate the utility of our EDU-based Context Compressor across two distinct scenarios: standard long-context benchmarks and complex open-domain Deep Search.

\subsubsection{General Long-Context Understanding}
We evaluate our LingEDU on {LongBench}~\citep{bai-etal-2024-longbench}, a benchmark covering multi-document QA, summarization, and few-shot learning. 

\noindent
\textbf{Baselines.}
We compare three configurations: (1) {Standard}, which feeds the full original context directly to the LLM; (2) {Self-Sum}, which utilizes the LLM itself to generate abstractive summaries of the context prior to processing; (3) {Glyph}~\citep{cheng2025glyph}, which applies an implicit compression baseline to serve as a reference for latent-based methods; and (4) {C3}~\citep{liu2025context}, which employs a small LLM to aggressively compress long contexts into compact latent representations before decoding with a LLM.

\paragraph{Analysis of Results.}
Table~\ref{tab:main_results} compares performance across Gemini-2.5-Pro and GPT-4.1 backbones. Overall, our EDU-based approach yields consistent improvements over the Standard baseline, with relative gains ($\Delta$) peaking at \textbf{+23.66\%} on few-shot tasks. Regarding Multi-Document QA (HotpotQA, Musique), our method outperforms Self-Sum. We attribute this to the tendency of abstractive methods to lose critical entities required for multi-hop reasoning; in contrast, our explicit tree structure preserves precise evidence chains via original text indices, enabling the model to look up exact details without hallucination. As for Summarization Tasks, while Self-Sum is naturally strong, our method remains competitive (e.g., surpassing Self-Sum on MultiNews by +1.44 points). Notably, it significantly outperforms the external Glyph baseline, suggesting that retaining hierarchical structure is more effective than latent compression for information retention.

\paragraph{Ablation Studies on Node Ranking Strategies.}
To isolate the impact of the ranking mechanism within LingoEDU, we decouple node selection from the reasoning backbone. We fix the generator as \textbf{GPT-4.1} across all experiments and compare five configurations:
(1) \textbf{Standard:} Feeds the full (or truncated) context directly without explicit filtering.
(2) \textbf{Random:} Randomly selects nodes to match our compression budget, serving as a stochastic lower bound.
(3) \textbf{BM25:} A sparse retrieval baseline relying on lexical overlap.
(4) \textbf{Self-Sum:} Prompts the generator (GPT-4.1) itself to identify relevant nodes prior to reasoning.
(5) \textbf{Ours (Qwen3-Reranker):} Semantically scores nodes using our lightweight Qwen3-Reranker-0.6B.

Table~\ref{tab:ablation_ranking} demonstrates that our dedicated ranking approach consistently outperforms other strategies. While the \textbf{Standard} baseline is hindered by noise in long contexts, structured selection significantly boosts performance. Crucially, \textbf{Ours} surpasses \textbf{Self-Sum} across most datasets (e.g., +2.2\% on HotpotQA, +1.83\% on DuReader). This result underscores that a specialized, lightweight dense ranker (0.6B)—despite its size—offers superior evidence localization compared to the intrinsic selection capabilities of a general-purpose LLM, which often lacks the granularity for precise context pruning. Furthermore, the substantial margin over \textbf{BM25} validates the necessity of semantic-aware filtering over surface-level matching.

\begin{table*}[t]
\centering
\resizebox{\textwidth}{!}{%
\begin{tabular}{l|cccc|cccc|cccc}
\toprule
\multirow{2}{*}{\textbf{Selection Strategy}} & \multicolumn{4}{c|}{\textbf{Multi-Doc QA}} & \multicolumn{4}{c|}{\textbf{Summarization}} & \multicolumn{4}{c}{\textbf{Few-shot}} \\
 & HotpotQA & 2Wiki & Musique & DuReader & GovRep & QMSum & MultiN & VCSum & TREC & Trivia & SAMSum & LSHT \\ \midrule
\textbf{Standard} (No Selection) & 65.83 & 72.98 & 51.90 & 21.80 & 29.97 & 22.84 & 20.85 & 12.50 & 77.00 & 90.07 & 39.20 & 48.60 \\
\textbf{Random} & 56.42 & 45.11 & 37.71 & 21.70 & 30.52 & 20.85 & 22.38 & 12.53 & 31.50 & 91.81 & 38.13 & 27.50 \\
\textbf{BM25} & 65.99 & 59.91 & 48.71 & \textbf{26.84} & 31.43 & 23.16 & 23.05 & 12.59 & 53.00 & 91.58 & 37.65 & 36.75 \\ \midrule
\textbf{Self-Sum} (LLM-Select) & \underline{67.89} & \underline{74.39} & \underline{53.48} & 23.51 & \underline{30.98} & \underline{22.53} & {22.06} & \underline{13.71} & {79.00} & \underline{93.69} & \underline{40.79} & \underline{50.50} \\ 
\textbf{Ours} (Qwen3-Reranker 0.6B) & \textbf{70.11} & \textbf{74.68} & \textbf{54.86} & 25.34 & \textbf{31.56} & \textbf{23.30} & \textbf{23.50} & \textbf{14.62} & \underline{80.00} & \textbf{93.76} & \textbf{41.68} & \textbf{52.50} \\ \bottomrule
\end{tabular}%
}
\caption{{Ablation study on ranking models.} All methods use GPT-4.1 as the final generator. We compare No Selection, Random, BM25, LLM Self-Selection, and Our Dedicated Reranker. }
\label{tab:ablation_ranking}
\end{table*}

\begin{table*}[h]
\centering
\resizebox{0.9\textwidth}{!}{%
\begin{tabular}{l|cccc|cccc}
\toprule
\multirow{2}{*}{\textbf{Model Backbone}} & \multicolumn{4}{c|}{\textbf{HLE}} & \multicolumn{4}{c}{\textbf{BrowseComp-ZH}} \\
\cmidrule(lr){2-5} \cmidrule(lr){6-9}
 & Base & Self-Sum & \textbf{Ours (LingoEDU)} & $\Delta$ & Base & Self-Sum & \textbf{Ours (LingoEDU)} & $\Delta$ \\
\midrule
DeepSeek-R1           & 9.0  & 9.5  & \textbf{13.6} & +51.11\% & 18.7 & 19.4 & \textbf{20.4} & +9.09\% \\
Qwen3-235B-Thinking   & 14.2 & 14.7 & \textbf{15.5} & +9.15\%  & 8.7  & 9.0  & \textbf{12.8} & +47.13\% \\
DeepSeek-V3.1         & 14.5 & 14.8 & \textbf{15.6} & +7.59\%  & 29.1 & 29.8 & \textbf{38.8} & +33.33\% \\
DeepSeek-V3.2         & 20.0 & 20.6 & \textbf{21.2} & +6.00\%  & 31.1  & 32.2 & \textbf{34.6} & +11.25\% \\
\midrule
\textit{Closed-Source Models} & & & & & & & & \\
GPT-5                 & 25.0 & 25.9 & \textbf{27.1} & +8.40\%  & 29.1 & 29.8 & \textbf{31.8} & +9.28\% \\
Claude Opus 4.1       & 14.0 & 14.8 & \textbf{15.5} & +10.71\% & 20.8 & 21.5 & \textbf{23.2} & +11.54\% \\
Gemini 3 Pro (w/o Deep Think)          & 26.1 & 26.7 & \textbf{30.1} & +15.33\% & 47.4 & 48.1 & \textbf{48.8} & +2.95\% \\
\bottomrule
\end{tabular}%
}
\caption{Ablation study of the LingoEDU module on {Deep Search}. Accuracy scores (\%) are reported. {Base}: Standard Deep Search without compression; {Self-Sum}: Query-focused summarization; {Ours (LingoEDU)}: Structural decomposition. $\Delta$ denotes the relative improvement of Ours over the Base baseline.}
\label{tab:edu_ablation}
\end{table*}

We further validate the effectiveness and efficiency of our framework through extensive supplementary analyses in Appendix~\ref{sec:Experiment_appendix}.

\subsubsection{Impact on Deep Search}

Unlike standard retrieval, {Deep Search} involves aggregating information from diverse, often noisy web sources to answer complex, open-ended queries. We integrated the LingoEDU module into a Deep Search pipeline and evaluated it on two challenging benchmarks(HLE~\citep{phan2025humanity} and BrowseComp-ZH~\citep{zhou2025browsecomp}) designed to push the limits of current LLMs.

\paragraph{Robustness to Web Noise.}
Deep search engines frequently ingest raw web pages cluttered with ads, navigation bars, and irrelevant links. As shown in Table~\ref{tab:edu_ablation}, on the noise-intensive \textit{BrowseComp-ZH}, our method delivers dramatic gains. Specifically, Qwen3-235B and DeepSeek-V3.1 show relative improvements of \textbf{47.13\%} and \textbf{33.33\%}, respectively.
This confirms that our Decomposer acts as a {semantic filter}: by identifying logical EDUs, it effectively prunes ``structural dead branches'' while preserving the core content, a capability that standard summarization lacks in such high-entropy Chinese web contexts.

\paragraph{Scaffolding for Complex Reasoning.}
 For the academic reasoning required by \textit{HLE}, where answers depend on synthesizing multiple distant clues across disciplines, DeepSeek-R1 gains \textbf{+51.11\%} relatively (from 9.0 to 13.6). This suggests that providing a cleaner, structure-aware context acts as a ``reasoning scaffold.'' It prevents the model from getting lost in irrelevant details (context drift), allowing it to focus its compute budget on logical deduction across valid evidence.

\paragraph{Compatibility with Frontier Models.}
Critically, our method remains additive even for the most advanced models like GPT-5 and Gemini 3 Pro (w/o Deep Think). This indicates that even as model capacity grows, handling unstructured noise and long-context reasoning remain bottlenecks. Our explicit compression provides a complementary signal that enhances the internal processing of SOTA LLMs.

\section{Related Work}

\subsection{Context Compression}
As the input context length for LLMs grows, compressing long documents into efficient representations has become a critical challenge. Existing approaches can generally be categorized into semantic summarization, soft prompting, and token-level pruning.

\paragraph{Explicit Compression Methods.}
Dominant approaches in this category operate on discrete tokens, filtering out low-informative content to reduce computational overhead.
Early methods like Selective-Context \cite{li:2023selective} and LLMLingua \cite{jiang2023:llmlingua} utilize perplexity-based metrics to prune redundant tokens.
Recent advances, such as LLMLingua-2 \cite{pan:2024llmlingua2} and TokenSkip \cite{xia2025tokenskip}, move towards data distillation and controllable pruning for higher efficiency.
While effective at reducing sequence length, these methods often operate on discrete tokens or rigid sentence boundaries, disrupting the local coherence of the text. 
Meanwhile, they typically focus on preserving the most important global information, while overlooking the original article’s structural information and fine-grained details.

\paragraph{Implicit Compression Methods.}
Alternatively, implicit methods map long contexts into continuous vector spaces or latent states.
Works like AutoCompressor \citep{chevalier-etal-2023-adapting} and ICAE \citep{ge2023context} compress text segments into soft prompts or memory slots.
More extreme approaches, such as 500xCompressor \citep{li2025500xcompressor} and Coconut \cite{hao2024training}, push this further by performing reasoning directly in the latent space.
However, recent studies \citep{li2025admtreecompressinglengthycontext} show that implicit compression methods tend to have positional bias. 
This means they often ignore information from the beginning or middle of the context, focusing instead on the most noticeable content and overlooking less prominent details.
Also, these implicit methods \citep{cheng2025glyphscalingcontextwindows, wei2025deepseekocrcontextsopticalcompression} tend to lack flexibility, as they often require specially designed post-training processes or the use of latent vectors as new inputs. 
This limits the applicability of such techniques to advanced API-based models.


Unlike these methods, our design allows our method to preserve global structure while capturing fine-grained details, and crucially, since it operates without latent representations, it ensures seamless compatibility with API-based models.

\subsection{Hallucinations in LLMs}
Hallucination in LLMs remains a pervasive issue, characterized by the generation of non-factual or unfaithful content \citep{hit-survey, si2025teaching}. 
Recent research has shifted from simply viewing hallucination as a generation error to a more nuanced perspective of ``controllable generation.''
A substantial body of work has taxonomized the causes of hallucination \cite{si2025aligning, liu2025prosocial, jiSurveyHallucinationNatural2023,zhao-etal-2025-looking}, treating hallucination mitigation as a debiasing task and striving to eliminate LLM hallucinations.
However, strictly eliminating such uncertainty may compromise model creativity and usability. \cite{jiangSurveyLargeLanguage2024} propose an alternative view, regarding hallucination as a manifestation of creativity that requires control rather than simply eliminating it. 
In this way, to mitigate the hallucination, many works attempt to introduce external knowledge integration to establish controllable context for the generation process of LLMs, e.g., RAG technologies.
However, these methods \citep{zhangHallucinationMitigationRetrievalAugmented2025, fanSurveyRAGMeeting2024} often rely on embedding-based retrieval stage and struggle with noisy retrieval or context integration due to lack of the structural relations of the retrieved context.
Different from these studies, we attempt to establish context for controllable generation through structured context compression.
Thus, our structured context compression via the proposed EDU-based Context Compressor can preserve global foresight while capturing fine-grained details, ensuring less noisy information and reducing the hallucination on downstream tasks.

\section{Conclusion}
In this work, we present the EDU-based Context Compressor, a plug-and-play framework bridging the gap between extended context windows and effective reasoning. By pivoting from linear reduction to hierarchical, discourse-aware decomposition, our method effectively filters noise while preserving the evidence chains required for complex tasks. We further validate this approach via StructBench, revealing that even state-of-the-art generalist LLMs lack the fine-grained structural analysis capabilities of our specialized model. Extensive experiments demonstrate that our method outperforms compression baselines and acts as a robust reasoning scaffold for search agents in noisy environments. These findings identify explicit structural modeling as a critical prerequisite for advanced long-context understanding. Future work will extend this paradigm to multi-modal contexts and dynamic agent memory.

\bibliography{biblio}

@article{ted_zhang1989simple,
  title = {Simple fast algorithms for the editing distance between trees and related problems},
  author = {Zhang, Kaizhong and Shasha, Dennis},
  journal = {SIAM journal on computing},
  volume = {18},
  number = {6},
  pages = {1245--1262},
  year = {1989},
  publisher = {SIAM}
}

@article{mann1988rhetorical,
  title={Rhetorical structure theory: Toward a functional theory of text organization},
  author={Mann, William C and Thompson, Sandra A},
  journal={Text-interdisciplinary Journal for the Study of Discourse},
  volume={8},
  number={3},
  pages={243--281},
  year={1988},
  publisher={Walter de Gruyter, Berlin/New York Berlin, New York}
}

@misc{wei2025deepseekocrcontextsopticalcompression,
      title={DeepSeek-OCR: Contexts Optical Compression}, 
      author={Haoran Wei and Yaofeng Sun and Yukun Li},
      year={2025},
      eprint={2510.18234},
      archivePrefix={arXiv},
      primaryClass={cs.CV},
      url={https://arxiv.org/abs/2510.18234}, 
}

@misc{cheng2025glyphscalingcontextwindows,
      title={Glyph: Scaling Context Windows via Visual-Text Compression}, 
      author={Jiale Cheng and Yusen Liu and Xinyu Zhang and Yulin Fei and Wenyi Hong and Ruiliang Lyu and Weihan Wang and Zhe Su and Xiaotao Gu and Xiao Liu and Yushi Bai and Jie Tang and Hongning Wang and Minlie Huang},
      year={2025},
      eprint={2510.17800},
      archivePrefix={arXiv},
      primaryClass={cs.CV},
      url={https://arxiv.org/abs/2510.17800}, 
}

@misc{li2025admtreecompressinglengthycontext,
      title={AdmTree: Compressing Lengthy Context with Adaptive Semantic Trees}, 
      author={Yangning Li and Shaoshen Chen and Yinghui Li and Yankai Chen and Hai-Tao Zheng and Hui Wang and Wenhao Jiang and Philip S. Yu},
      year={2025},
      eprint={2512.04550},
      archivePrefix={arXiv},
      primaryClass={cs.CL},
      url={https://arxiv.org/abs/2512.04550}, 
}

@misc{liu2025contextcascadecompressionexploring,
      title={Context Cascade Compression: Exploring the Upper Limits of Text Compression}, 
      author={Fanfan Liu and Haibo Qiu},
      year={2025},
      eprint={2511.15244},
      archivePrefix={arXiv},
      primaryClass={cs.CL},
      url={https://arxiv.org/abs/2511.15244}, 
}

@misc{xu2023recompimprovingretrievalaugmentedlms,
      title={RECOMP: Improving Retrieval-Augmented LMs with Compression and Selective Augmentation}, 
      author={Fangyuan Xu and Weijia Shi and Eunsol Choi},
      year={2023},
      eprint={2310.04408},
      archivePrefix={arXiv},
      primaryClass={cs.CL},
      url={https://arxiv.org/abs/2310.04408}, 
}

@misc{hua2025contextengineering20context,
      title={Context Engineering 2.0: The Context of Context Engineering}, 
      author={Qishuo Hua and Lyumanshan Ye and Dayuan Fu and Yang Xiao and Xiaojie Cai and Yunze Wu and Jifan Lin and Junfei Wang and Pengfei Liu},
      year={2025},
      eprint={2510.26493},
      archivePrefix={arXiv},
      primaryClass={cs.AI},
      url={https://arxiv.org/abs/2510.26493}, 
}

@misc{kryściński2022booksumcollectiondatasetslongform,
      title={BookSum: A Collection of Datasets for Long-form Narrative Summarization}, 
      author={Wojciech Kryściński and Nazneen Rajani and Divyansh Agarwal and Caiming Xiong and Dragomir Radev},
      year={2022},
      eprint={2105.08209},
      archivePrefix={arXiv},
      primaryClass={cs.CL},
      url={https://arxiv.org/abs/2105.08209}, 
}

@inproceedings{bai-etal-2024-longbench,
    title = "{L}ong{B}ench: A Bilingual, Multitask Benchmark for Long Context Understanding",
    author = "Bai, Yushi  and
      Lv, Xin  and
      Zhang, Jiajie  and
      Lyu, Hongchang  and
      Tang, Jiankai  and
      Huang, Zhidian  and
      Du, Zhengxiao  and
      Liu, Xiao  and
      Zeng, Aohan  and
      Hou, Lei  and
      Dong, Yuxiao  and
      Tang, Jie  and
      Li, Juanzi",
    editor = "Ku, Lun-Wei  and
      Martins, Andre  and
      Srikumar, Vivek",
    booktitle = "Proceedings of the 62nd Annual Meeting of the Association for Computational Linguistics (Volume 1: Long Papers)",
    month = aug,
    year = "2024",
    address = "Bangkok, Thailand",
    publisher = "Association for Computational Linguistics",
    url = "https://aclanthology.org/2024.acl-long.172/",
    doi = "10.18653/v1/2024.acl-long.172",
    pages = "3119--3137"
}

@article{luo2025large,
  title = {Large language models surpass human experts in predicting neuroscience results},
  author = {Luo, Xiaoliang and Rechardt, Akilles and Sun, Guangzhi and Nejad, Kevin K and Y{\'a}{\~n}ez, Felipe and Yilmaz, Bati and Lee, Kangjoo and Cohen, Alexandra O and Borghesani, Valentina and Pashkov, Anton and others},
  journal = {Nature human behaviour},
  volume = {9},
  number = {2},
  pages = {305--315},
  year = {2025},
  publisher = {Nature Publishing Group UK London}
}

@article{gpt-4o,
  title={Gpt-4o system card},
  author={Hurst, Aaron and Lerer, Adam and Goucher, Adam P and Perelman, Adam and Ramesh, Aditya and Clark, Aidan and Ostrow, AJ and Welihinda, Akila and Hayes, Alan and Radford, Alec and others},
  journal={arXiv preprint arXiv:2410.21276},
  year={2024}
}

@inproceedings{an-etal-2025-ultraif,
    title = "{U}ltra{IF}: Advancing Instruction Following from the Wild",
    author = "An, Kaikai  and
      Sheng, Li  and
      Cui, Ganqu  and
      Si, Shuzheng  and
      Ding, Ning  and
      Cheng, Yu  and
      Chang, Baobao",
    editor = "Christodoulopoulos, Christos  and
      Chakraborty, Tanmoy  and
      Rose, Carolyn  and
      Peng, Violet",
    booktitle = "Proceedings of the 2025 Conference on Empirical Methods in Natural Language Processing",
    month = nov,
    year = "2025",
    address = "Suzhou, China",
    publisher = "Association for Computational Linguistics",
    url = "https://aclanthology.org/2025.emnlp-main.945/",
    doi = "10.18653/v1/2025.emnlp-main.945",
    pages = "18722--18737",
    ISBN = "979-8-89176-332-6"
}

@inproceedings{zhao-etal-2025-looking,
    title = "Looking Beyond Text: Reducing Language Bias in Large Vision-Language Models via Multimodal Dual-Attention and Soft-Image Guidance",
    author = "Zhao, Haozhe  and
      Si, Shuzheng  and
      Chen, Liang  and
      Zhang, Yichi  and
      Sun, Maosong  and
      Chang, Baobao  and
      Zhang, Minjia",
    editor = "Christodoulopoulos, Christos  and
      Chakraborty, Tanmoy  and
      Rose, Carolyn  and
      Peng, Violet",
    booktitle = "Proceedings of the 2025 Conference on Empirical Methods in Natural Language Processing",
    month = nov,
    year = "2025",
    address = "Suzhou, China",
    publisher = "Association for Computational Linguistics",
    url = "https://aclanthology.org/2025.emnlp-main.995/",
    doi = "10.18653/v1/2025.emnlp-main.995",
    pages = "19677--19701",
    ISBN = "979-8-89176-332-6"
}

@inproceedings{wang-etal-2025-document,
    title = "Document Segmentation Matters for Retrieval-Augmented Generation",
    author = "Wang, Zhitong  and
      Gao, Cheng  and
      Xiao, Chaojun  and
      Huang, Yufei  and
      Si, Shuzheng  and
      Luo, Kangyang  and
      Bai, Yuzhuo  and
      Li, Wenhao  and
      Duan, Tangjian  and
      Lv, Chuancheng  and
      Lu, Guoshan  and
      Chen, Gang  and
      Qi, Fanchao  and
      Sun, Maosong",
    editor = "Che, Wanxiang  and
      Nabende, Joyce  and
      Shutova, Ekaterina  and
      Pilehvar, Mohammad Taher",
    booktitle = "Findings of the Association for Computational Linguistics: ACL 2025",
    month = jul,
    year = "2025",
    address = "Vienna, Austria",
    publisher = "Association for Computational Linguistics",
    url = "https://aclanthology.org/2025.findings-acl.422/",
    doi = "10.18653/v1/2025.findings-acl.422",
    pages = "8063--8075",
    ISBN = "979-8-89176-256-5"
}

@inproceedings{si-etal-2025-gateau,
    title = "{GATEAU}: Selecting Influential Samples for Long Context Alignment",
    author = "Si, Shuzheng  and
      Zhao, Haozhe  and
      Chen, Gang  and
      Li, Yunshui  and
      Luo, Kangyang  and
      Lv, Chuancheng  and
      An, Kaikai  and
      Qi, Fanchao  and
      Chang, Baobao  and
      Sun, Maosong",
    editor = "Christodoulopoulos, Christos  and
      Chakraborty, Tanmoy  and
      Rose, Carolyn  and
      Peng, Violet",
    booktitle = "Proceedings of the 2025 Conference on Empirical Methods in Natural Language Processing",
    month = nov,
    year = "2025",
    address = "Suzhou, China",
    publisher = "Association for Computational Linguistics",
    url = "https://aclanthology.org/2025.emnlp-main.375/",
    doi = "10.18653/v1/2025.emnlp-main.375",
    pages = "7391--7422",
    ISBN = "979-8-89176-332-6"
}

@article{si2025teaching,
  title={Teaching Large Language Models to Maintain Contextual Faithfulness via Synthetic Tasks and Reinforcement Learning},
  author={Si, Shuzheng and Zhao, Haozhe and Gao, Cheng and Bai, Yuzhuo and Wang, Zhitong and Gao, Bofei and Luo, Kangyang and Li, Wenhao and Huang, Yufei and Chen, Gang and others},
  journal={arXiv preprint arXiv:2505.16483},
  year={2025}
}

@article{yi2024survey,
  title={A survey on recent advances in llm-based multi-turn dialogue systems},
  author={Yi, Zihao and Ouyang, Jiarui and Liu, Yuwen and Liao, Tianhao and Xu, Zhe and Shen, Ying},
  journal={arXiv preprint arXiv:2402.18013},
  year={2024}
}

@article{qwen3,
    title={Qwen3 Technical Report}, 
    author={An Yang and Anfeng Li and Baosong Yang and Beichen Zhang and Binyuan Hui and Bo Zheng and Bowen Yu and Chang Gao and Chengen Huang and Chenxu Lv and Chujie Zheng and Dayiheng Liu and Fan Zhou and Fei Huang and Feng Hu and Hao Ge and Haoran Wei and Huan Lin and Jialong Tang and Jian Yang and Jianhong Tu and Jianwei Zhang and Jianxin Yang and Jiaxi Yang and Jing Zhou and Jingren Zhou and Junyang Lin and Kai Dang and Keqin Bao and Kexin Yang and Le Yu and Lianghao Deng and Mei Li and Mingfeng Xue and Mingze Li and Pei Zhang and Peng Wang and Qin Zhu and Rui Men and Ruize Gao and Shixuan Liu and Shuang Luo and Tianhao Li and Tianyi Tang and Wenbiao Yin and Xingzhang Ren and Xinyu Wang and Xinyu Zhang and Xuancheng Ren and Yang Fan and Yang Su and Yichang Zhang and Yinger Zhang and Yu Wan and Yuqiong Liu and Zekun Wang and Zeyu Cui and Zhenru Zhang and Zhipeng Zhou and Zihan Qiu},
    journal = {arXiv preprint arXiv:2505.09388},
    year={2025}
}

@misc{deepseekai2025deepseekv3technicalreport,
      title={DeepSeek-V3 Technical Report}, 
      author={DeepSeek-AI and Aixin Liu and Bei Feng and Bing Xue and Bingxuan Wang and Bochao Wu and Chengda Lu and Chenggang Zhao and Chengqi Deng and Chenyu Zhang and Chong Ruan and Damai Dai and Daya Guo and Dejian Yang and Deli Chen and Dongjie Ji and Erhang Li and Fangyun Lin and Fucong Dai and Fuli Luo and Guangbo Hao and Guanting Chen and Guowei Li and H. Zhang and Han Bao and Hanwei Xu and Haocheng Wang and Haowei Zhang and Honghui Ding and Huajian Xin and Huazuo Gao and Hui Li and Hui Qu and J. L. Cai and Jian Liang and Jianzhong Guo and Jiaqi Ni and Jiashi Li and Jiawei Wang and Jin Chen and Jingchang Chen and Jingyang Yuan and Junjie Qiu and Junlong Li and Junxiao Song and Kai Dong and Kai Hu and Kaige Gao and Kang Guan and Kexin Huang and Kuai Yu and Lean Wang and Lecong Zhang and Lei Xu and Leyi Xia and Liang Zhao and Litong Wang and Liyue Zhang and Meng Li and Miaojun Wang and Mingchuan Zhang and Minghua Zhang and Minghui Tang and Mingming Li and Ning Tian and Panpan Huang and Peiyi Wang and Peng Zhang and Qiancheng Wang and Qihao Zhu and Qinyu Chen and Qiushi Du and R. J. Chen and R. L. Jin and Ruiqi Ge and Ruisong Zhang and Ruizhe Pan and Runji Wang and Runxin Xu and Ruoyu Zhang and Ruyi Chen and S. S. Li and Shanghao Lu and Shangyan Zhou and Shanhuang Chen and Shaoqing Wu and Shengfeng Ye and Shengfeng Ye and Shirong Ma and Shiyu Wang and Shuang Zhou and Shuiping Yu and Shunfeng Zhou and Shuting Pan and T. Wang and Tao Yun and Tian Pei and Tianyu Sun and W. L. Xiao and Wangding Zeng and Wanjia Zhao and Wei An and Wen Liu and Wenfeng Liang and Wenjun Gao and Wenqin Yu and Wentao Zhang and X. Q. Li and Xiangyue Jin and Xianzu Wang and Xiao Bi and Xiaodong Liu and Xiaohan Wang and Xiaojin Shen and Xiaokang Chen and Xiaokang Zhang and Xiaosha Chen and Xiaotao Nie and Xiaowen Sun and Xiaoxiang Wang and Xin Cheng and Xin Liu and Xin Xie and Xingchao Liu and Xingkai Yu and Xinnan Song and Xinxia Shan and Xinyi Zhou and Xinyu Yang and Xinyuan Li and Xuecheng Su and Xuheng Lin and Y. K. Li and Y. Q. Wang and Y. X. Wei and Y. X. Zhu and Yang Zhang and Yanhong Xu and Yanhong Xu and Yanping Huang and Yao Li and Yao Zhao and Yaofeng Sun and Yaohui Li and Yaohui Wang and Yi Yu and Yi Zheng and Yichao Zhang and Yifan Shi and Yiliang Xiong and Ying He and Ying Tang and Yishi Piao and Yisong Wang and Yixuan Tan and Yiyang Ma and Yiyuan Liu and Yongqiang Guo and Yu Wu and Yuan Ou and Yuchen Zhu and Yuduan Wang and Yue Gong and Yuheng Zou and Yujia He and Yukun Zha and Yunfan Xiong and Yunxian Ma and Yuting Yan and Yuxiang Luo and Yuxiang You and Yuxuan Liu and Yuyang Zhou and Z. F. Wu and Z. Z. Ren and Zehui Ren and Zhangli Sha and Zhe Fu and Zhean Xu and Zhen Huang and Zhen Zhang and Zhenda Xie and Zhengyan Zhang and Zhewen Hao and Zhibin Gou and Zhicheng Ma and Zhigang Yan and Zhihong Shao and Zhipeng Xu and Zhiyu Wu and Zhongyu Zhang and Zhuoshu Li and Zihui Gu and Zijia Zhu and Zijun Liu and Zilin Li and Ziwei Xie and Ziyang Song and Ziyi Gao and Zizheng Pan},
      year={2025},
      eprint={2412.19437},
      archivePrefix={arXiv},
      primaryClass={cs.CL},
      url={https://arxiv.org/abs/2412.19437}, 
}

@misc{Claude4,
  title={Introducing Claude 4},
  author={Anthropic},
  year={2025},
  url={https://www.anthropic.com/news/claude-4}
}

@misc{gpt-4.1,
  title={Introducing GPT-4.1 in the API},
  author={OpenAI},
  year={2025},
  url={https://openai.com/index/gpt-4-1/}
}

@misc{o3,
  title={OpenAI o3 and o4-mini System Card},
  author={OpenAI},
  year={2025},
  url={https://cdn.openai.com/pdf/2221c875-02dc-4789-800b-e7758f3722c1/o3-and-o4-mini-system-card.pdf}
}

@article{si2025aligning,
  title={Aligning Large Language Models to Follow Instructions and Hallucinate Less via Effective Data Filtering},
  author={Si, Shuzheng and Zhao, Haozhe and Chen, Gang and Gao, Cheng and Bai, Yuzhuo and Wang, Zhitong and An, Kaikai and Luo, Kangyang and Qian, Chen and Qi, Fanchao and others},
  journal={arXiv preprint arXiv:2502.07340},
  year={2025}
}

@article{chen2021dialogsum,
  title={DialogSum: A real-life scenario dialogue summarization dataset},
  author={Chen, Yulong and Liu, Yang and Chen, Liang and Zhang, Yue},
  journal={arXiv preprint arXiv:2105.06762},
  year={2021}
}

@inproceedings{
si2023spokenwoz,
title={Spoken{WOZ}: A Large-Scale Speech-Text Benchmark for Spoken Task-Oriented Dialogue Agents},
author={Shuzheng Si and Wentao Ma and Haoyu Gao and Yuchuan Wu and Ting-En Lin and Yinpei Dai and Hangyu Li and Rui Yan and Fei Huang and Yongbin Li},
booktitle={Thirty-seventh Conference on Neural Information Processing Systems Datasets and Benchmarks Track},
year={2023},
url={https://openreview.net/forum?id=viktK3nO5b}
}

@article{bai2023longbench,
  title={LongBench: A Bilingual, Multitask Benchmark for Long Context Understanding},
  author={Bai, Yushi and Lv, Xin and Zhang, Jiajie and Lyu, Hongchang and Tang, Jiankai and Huang, Zhidian and Du, Zhengxiao and Liu, Xiao and Zeng, Aohan and Hou, Lei and Dong, Yuxiao and Tang, Jie and Li, Juanzi},
  journal={arXiv preprint arXiv:2308.14508},
  year={2023}
}

@inproceedings{si-etal-2022-scl,
    title = "{SCL}-{RAI}: Span-based Contrastive Learning with Retrieval Augmented Inference for Unlabeled Entity Problem in {NER}",
    author = "Si, Shuzheng  and
      Zeng, Shuang  and
      Lin, Jiaxing  and
      Chang, Baobao",
    editor = "Calzolari, Nicoletta  and
      Huang, Chu-Ren  and
      Kim, Hansaem  and
      Pustejovsky, James  and
      Wanner, Leo  and
      Choi, Key-Sun  and
      Ryu, Pum-Mo  and
      Chen, Hsin-Hsi  and
      Donatelli, Lucia  and
      Ji, Heng  and
      Kurohashi, Sadao  and
      Paggio, Patrizia  and
      Xue, Nianwen  and
      Kim, Seokhwan  and
      Hahm, Younggyun  and
      He, Zhong  and
      Lee, Tony Kyungil  and
      Santus, Enrico  and
      Bond, Francis  and
      Na, Seung-Hoon",
    booktitle = "Proceedings of the 29th International Conference on Computational Linguistics",
    month = oct,
    year = "2022",
    address = "Gyeongju, Republic of Korea",
    publisher = "International Committee on Computational Linguistics",
    url = "https://aclanthology.org/2022.coling-1.202",
    pages = "2313--2318",
}

@misc{si2025goalplanjustwish,
      title={A Goal Without a Plan Is Just a Wish: Efficient and Effective Global Planner Training for Long-Horizon Agent Tasks}, 
      author={Shuzheng Si and Haozhe Zhao and Kangyang Luo and Gang Chen and Fanchao Qi and Minjia Zhang and Baobao Chang and Maosong Sun},
      year={2025},
      eprint={2510.05608},
      archivePrefix={arXiv},
      primaryClass={cs.CL},
      url={https://arxiv.org/abs/2510.05608}, 
}

@misc{llama3,
      title={The Llama 3 Herd of Models}, 
      author={Aaron Grattafiori and Abhimanyu Dubey and Abhinav Jauhri and Abhinav Pandey and Abhishek Kadian and Ahmad Al-Dahle and Aiesha Letman and Akhil Mathur and Alan Schelten and Alex Vaughan and Amy Yang and Angela Fan and Anirudh Goyal and Anthony Hartshorn and Aobo Yang and Archi Mitra and Archie Sravankumar and Artem Korenev and Arthur Hinsvark and Arun Rao and Aston Zhang and Aurelien Rodriguez and Austen Gregerson and Ava Spataru and Baptiste Roziere and Bethany Biron and Binh Tang and Bobbie Chern and Charlotte Caucheteux and Chaya Nayak and Chloe Bi and Chris Marra and Chris McConnell and Christian Keller and Christophe Touret and Chunyang Wu and Corinne Wong and Cristian Canton Ferrer and Cyrus Nikolaidis and Damien Allonsius and Daniel Song and Danielle Pintz and Danny Livshits and Danny Wyatt and David Esiobu and Dhruv Choudhary and Dhruv Mahajan and Diego Garcia-Olano and Diego Perino and Dieuwke Hupkes and Egor Lakomkin and Ehab AlBadawy and Elina Lobanova and Emily Dinan and Eric Michael Smith and Filip Radenovic and Francisco Guzmán and Frank Zhang and Gabriel Synnaeve and Gabrielle Lee and Georgia Lewis Anderson and Govind Thattai and Graeme Nail and Gregoire Mialon and Guan Pang and Guillem Cucurell and Hailey Nguyen and Hannah Korevaar and Hu Xu and Hugo Touvron and Iliyan Zarov and Imanol Arrieta Ibarra and Isabel Kloumann and Ishan Misra and Ivan Evtimov and Jack Zhang and Jade Copet and Jaewon Lee and Jan Geffert and Jana Vranes and Jason Park and Jay Mahadeokar and Jeet Shah and Jelmer van der Linde and Jennifer Billock and Jenny Hong and Jenya Lee and Jeremy Fu and Jianfeng Chi and Jianyu Huang and Jiawen Liu and Jie Wang and Jiecao Yu and Joanna Bitton and Joe Spisak and Jongsoo Park and Joseph Rocca and Joshua Johnstun and Joshua Saxe and Junteng Jia and Kalyan Vasuden Alwala and Karthik Prasad and Kartikeya Upasani and Kate Plawiak and Ke Li and Kenneth Heafield and Kevin Stone and Khalid El-Arini and Krithika Iyer and Kshitiz Malik and Kuenley Chiu and Kunal Bhalla and Kushal Lakhotia and Lauren Rantala-Yeary and Laurens van der Maaten and Lawrence Chen and Liang Tan and Liz Jenkins and Louis Martin and Lovish Madaan and Lubo Malo and Lukas Blecher and Lukas Landzaat and Luke de Oliveira and Madeline Muzzi and Mahesh Pasupuleti and Mannat Singh and Manohar Paluri and Marcin Kardas and Maria Tsimpoukelli and Mathew Oldham and Mathieu Rita and Maya Pavlova and Melanie Kambadur and Mike Lewis and Min Si and Mitesh Kumar Singh and Mona Hassan and Naman Goyal and Narjes Torabi and Nikolay Bashlykov and Nikolay Bogoychev and Niladri Chatterji and Ning Zhang and Olivier Duchenne and Onur Çelebi and Patrick Alrassy and Pengchuan Zhang and Pengwei Li and Petar Vasic and Peter Weng and Prajjwal Bhargava and Pratik Dubal and Praveen Krishnan and Punit Singh Koura and Puxin Xu and Qing He and Qingxiao Dong and Ragavan Srinivasan and Raj Ganapathy and Ramon Calderer and Ricardo Silveira Cabral and Robert Stojnic and Roberta Raileanu and Rohan Maheswari and Rohit Girdhar and Rohit Patel and Romain Sauvestre and Ronnie Polidoro and Roshan Sumbaly and Ross Taylor and Ruan Silva and Rui Hou and Rui Wang and Saghar Hosseini and Sahana Chennabasappa and Sanjay Singh and Sean Bell and Seohyun Sonia Kim and Sergey Edunov and Shaoliang Nie and Sharan Narang and Sharath Raparthy and Sheng Shen and Shengye Wan and Shruti Bhosale and Shun Zhang and Simon Vandenhende and Soumya Batra and Spencer Whitman and Sten Sootla and Stephane Collot and Suchin Gururangan and Sydney Borodinsky and Tamar Herman and Tara Fowler and Tarek Sheasha and Thomas Georgiou and Thomas Scialom and Tobias Speckbacher and Todor Mihaylov and Tong Xiao and Ujjwal Karn and Vedanuj Goswami and Vibhor Gupta and Vignesh Ramanathan and Viktor Kerkez and Vincent Gonguet and Virginie Do and Vish Vogeti and Vítor Albiero and Vladan Petrovic and Weiwei Chu and Wenhan Xiong and Wenyin Fu and Whitney Meers and Xavier Martinet and Xiaodong Wang and Xiaofang Wang and Xiaoqing Ellen Tan and Xide Xia and Xinfeng Xie and Xuchao Jia and Xuewei Wang and Yaelle Goldschlag and Yashesh Gaur and Yasmine Babaei and Yi Wen and Yiwen Song and Yuchen Zhang and Yue Li and Yuning Mao and Zacharie Delpierre Coudert and Zheng Yan and Zhengxing Chen and Zoe Papakipos and Aaditya Singh and Aayushi Srivastava and Abha Jain and Adam Kelsey and Adam Shajnfeld and Adithya Gangidi and Adolfo Victoria and Ahuva Goldstand and Ajay Menon and Ajay Sharma and Alex Boesenberg and Alexei Baevski and Allie Feinstein and Amanda Kallet and Amit Sangani and Amos Teo and Anam Yunus and Andrei Lupu and Andres Alvarado and Andrew Caples and Andrew Gu and Andrew Ho and Andrew Poulton and Andrew Ryan and Ankit Ramchandani and Annie Dong and Annie Franco and Anuj Goyal and Aparajita Saraf and Arkabandhu Chowdhury and Ashley Gabriel and Ashwin Bharambe and Assaf Eisenman and Azadeh Yazdan and Beau James and Ben Maurer and Benjamin Leonhardi and Bernie Huang and Beth Loyd and Beto De Paola and Bhargavi Paranjape and Bing Liu and Bo Wu and Boyu Ni and Braden Hancock and Bram Wasti and Brandon Spence and Brani Stojkovic and Brian Gamido and Britt Montalvo and Carl Parker and Carly Burton and Catalina Mejia and Ce Liu and Changhan Wang and Changkyu Kim and Chao Zhou and Chester Hu and Ching-Hsiang Chu and Chris Cai and Chris Tindal and Christoph Feichtenhofer and Cynthia Gao and Damon Civin and Dana Beaty and Daniel Kreymer and Daniel Li and David Adkins and David Xu and Davide Testuggine and Delia David and Devi Parikh and Diana Liskovich and Didem Foss and Dingkang Wang and Duc Le and Dustin Holland and Edward Dowling and Eissa Jamil and Elaine Montgomery and Eleonora Presani and Emily Hahn and Emily Wood and Eric-Tuan Le and Erik Brinkman and Esteban Arcaute and Evan Dunbar and Evan Smothers and Fei Sun and Felix Kreuk and Feng Tian and Filippos Kokkinos and Firat Ozgenel and Francesco Caggioni and Frank Kanayet and Frank Seide and Gabriela Medina Florez and Gabriella Schwarz and Gada Badeer and Georgia Swee and Gil Halpern and Grant Herman and Grigory Sizov and Guangyi and Zhang and Guna Lakshminarayanan and Hakan Inan and Hamid Shojanazeri and Han Zou and Hannah Wang and Hanwen Zha and Haroun Habeeb and Harrison Rudolph and Helen Suk and Henry Aspegren and Hunter Goldman and Hongyuan Zhan and Ibrahim Damlaj and Igor Molybog and Igor Tufanov and Ilias Leontiadis and Irina-Elena Veliche and Itai Gat and Jake Weissman and James Geboski and James Kohli and Janice Lam and Japhet Asher and Jean-Baptiste Gaya and Jeff Marcus and Jeff Tang and Jennifer Chan and Jenny Zhen and Jeremy Reizenstein and Jeremy Teboul and Jessica Zhong and Jian Jin and Jingyi Yang and Joe Cummings and Jon Carvill and Jon Shepard and Jonathan McPhie and Jonathan Torres and Josh Ginsburg and Junjie Wang and Kai Wu and Kam Hou U and Karan Saxena and Kartikay Khandelwal and Katayoun Zand and Kathy Matosich and Kaushik Veeraraghavan and Kelly Michelena and Keqian Li and Kiran Jagadeesh and Kun Huang and Kunal Chawla and Kyle Huang and Lailin Chen and Lakshya Garg and Lavender A and Leandro Silva and Lee Bell and Lei Zhang and Liangpeng Guo and Licheng Yu and Liron Moshkovich and Luca Wehrstedt and Madian Khabsa and Manav Avalani and Manish Bhatt and Martynas Mankus and Matan Hasson and Matthew Lennie and Matthias Reso and Maxim Groshev and Maxim Naumov and Maya Lathi and Meghan Keneally and Miao Liu and Michael L. Seltzer and Michal Valko and Michelle Restrepo and Mihir Patel and Mik Vyatskov and Mikayel Samvelyan and Mike Clark and Mike Macey and Mike Wang and Miquel Jubert Hermoso and Mo Metanat and Mohammad Rastegari and Munish Bansal and Nandhini Santhanam and Natascha Parks and Natasha White and Navyata Bawa and Nayan Singhal and Nick Egebo and Nicolas Usunier and Nikhil Mehta and Nikolay Pavlovich Laptev and Ning Dong and Norman Cheng and Oleg Chernoguz and Olivia Hart and Omkar Salpekar and Ozlem Kalinli and Parkin Kent and Parth Parekh and Paul Saab and Pavan Balaji and Pedro Rittner and Philip Bontrager and Pierre Roux and Piotr Dollar and Polina Zvyagina and Prashant Ratanchandani and Pritish Yuvraj and Qian Liang and Rachad Alao and Rachel Rodriguez and Rafi Ayub and Raghotham Murthy and Raghu Nayani and Rahul Mitra and Rangaprabhu Parthasarathy and Raymond Li and Rebekkah Hogan and Robin Battey and Rocky Wang and Russ Howes and Ruty Rinott and Sachin Mehta and Sachin Siby and Sai Jayesh Bondu and Samyak Datta and Sara Chugh and Sara Hunt and Sargun Dhillon and Sasha Sidorov and Satadru Pan and Saurabh Mahajan and Saurabh Verma and Seiji Yamamoto and Sharadh Ramaswamy and Shaun Lindsay and Shaun Lindsay and Sheng Feng and Shenghao Lin and Shengxin Cindy Zha and Shishir Patil and Shiva Shankar and Shuqiang Zhang and Shuqiang Zhang and Sinong Wang and Sneha Agarwal and Soji Sajuyigbe and Soumith Chintala and Stephanie Max and Stephen Chen and Steve Kehoe and Steve Satterfield and Sudarshan Govindaprasad and Sumit Gupta and Summer Deng and Sungmin Cho and Sunny Virk and Suraj Subramanian and Sy Choudhury and Sydney Goldman and Tal Remez and Tamar Glaser and Tamara Best and Thilo Koehler and Thomas Robinson and Tianhe Li and Tianjun Zhang and Tim Matthews and Timothy Chou and Tzook Shaked and Varun Vontimitta and Victoria Ajayi and Victoria Montanez and Vijai Mohan and Vinay Satish Kumar and Vishal Mangla and Vlad Ionescu and Vlad Poenaru and Vlad Tiberiu Mihailescu and Vladimir Ivanov and Wei Li and Wenchen Wang and Wenwen Jiang and Wes Bouaziz and Will Constable and Xiaocheng Tang and Xiaojian Wu and Xiaolan Wang and Xilun Wu and Xinbo Gao and Yaniv Kleinman and Yanjun Chen and Ye Hu and Ye Jia and Ye Qi and Yenda Li and Yilin Zhang and Ying Zhang and Yossi Adi and Youngjin Nam and Yu and Wang and Yu Zhao and Yuchen Hao and Yundi Qian and Yunlu Li and Yuzi He and Zach Rait and Zachary DeVito and Zef Rosnbrick and Zhaoduo Wen and Zhenyu Yang and Zhiwei Zhao and Zhiyu Ma},
      year={2024},
      eprint={2407.21783},
      archivePrefix={arXiv},
      primaryClass={cs.AI},
      url={https://arxiv.org/abs/2407.21783}, 
}

@article{hit-survey,
author = {Huang, Lei and Yu, Weijiang and Ma, Weitao and Zhong, Weihong and Feng, Zhangyin and Wang, Haotian and Chen, Qianglong and Peng, Weihua and Feng, Xiaocheng and Qin, Bing and Liu, Ting},
title = {A Survey on Hallucination in Large Language Models: Principles, Taxonomy, Challenges, and Open Questions},
year = {2024},
publisher = {Association for Computing Machinery},
address = {New York, NY, USA},
issn = {1046-8188},
url = {https://doi.org/10.1145/3703155},
doi = {10.1145/3703155},
note = {Just Accepted},
journal = {ACM Trans. Inf. Syst.},
month = nov
}

@article{gpt4,
  title={GPT-4 Technical Report},
  author={OpenAI},
  journal={ArXiv},
  year={2023},
  volume={abs/2303.08774}
}

@inproceedings{li:2023selective,
    title = "Compressing Context to Enhance Inference Efficiency of Large Language Models",
    author = "Li, Yucheng  and
      Dong, Bo  and
      Guerin, Frank  and
      Lin, Chenghua",
    editor = "Bouamor, Houda  and
      Pino, Juan  and
      Bali, Kalika",
    booktitle = "Proceedings of the 2023 Conference on Empirical Methods in Natural Language Processing",
    month = dec,
    year = "2023",
    address = "Singapore",
    publisher = "Association for Computational Linguistics",
    url = "https://aclanthology.org/2023.emnlp-main.391/",
    doi = "10.18653/v1/2023.emnlp-main.391",
    pages = "6342--6353",
}

@inproceedings{pan:2024llmlingua2,
    title = "{LLML}ingua-2: Data Distillation for Efficient and Faithful Task-Agnostic Prompt Compression",
    author = {Pan, Zhuoshi  and
      Wu, Qianhui  and
      Jiang, Huiqiang  and
      Xia, Menglin  and
      Luo, Xufang  and
      Zhang, Jue  and
      Lin, Qingwei  and
      R{\"u}hle, Victor  and
      Yang, Yuqing  and
      Lin, Chin-Yew  and
      Zhao, H. Vicky  and
      Qiu, Lili  and
      Zhang, Dongmei},
    editor = "Ku, Lun-Wei  and
      Martins, Andre  and
      Srikumar, Vivek",
    booktitle = "Findings of the Association for Computational Linguistics: ACL 2024",
    month = aug,
    year = "2024",
    address = "Bangkok, Thailand",
    publisher = "Association for Computational Linguistics",
    url = "https://aclanthology.org/2024.findings-acl.57/",
    doi = "10.18653/v1/2024.findings-acl.57",
    pages = "963--981",
}

@inproceedings{jiang2023:llmlingua,
    title = "{LLML}ingua: Compressing Prompts for Accelerated Inference of Large Language Models",
    author = "Jiang, Huiqiang  and
      Wu, Qianhui  and
      Lin, Chin-Yew  and
      Yang, Yuqing  and
      Qiu, Lili",
    editor = "Bouamor, Houda  and
      Pino, Juan  and
      Bali, Kalika",
    booktitle = "Proceedings of the 2023 Conference on Empirical Methods in Natural Language Processing",
    month = dec,
    year = "2023",
    address = "Singapore",
    publisher = "Association for Computational Linguistics",
    url = "https://aclanthology.org/2023.emnlp-main.825/",
    doi = "10.18653/v1/2023.emnlp-main.825",
    pages = "13358--13376",
}

@inproceedings{chevalier-etal-2023-adapting,
    title = "Adapting Language Models to Compress Contexts",
    author = "Chevalier, Alexis  and
      Wettig, Alexander  and
      Ajith, Anirudh  and
      Chen, Danqi",
    editor = "Bouamor, Houda  and
      Pino, Juan  and
      Bali, Kalika",
    booktitle = "Proceedings of the 2023 Conference on Empirical Methods in Natural Language Processing",
    month = dec,
    year = "2023",
    address = "Singapore",
    publisher = "Association for Computational Linguistics",
    url = "https://aclanthology.org/2023.emnlp-main.232/",
    doi = "10.18653/v1/2023.emnlp-main.232",
    pages = "3829--3846",
}

@article{jiSurveyHallucinationNatural2023,
 title = {Survey of Hallucination in Natural Language Generation},
 author = {Ji, Ziwei and Lee, Nayeon and Frieske, Rita and Yu, Tiezheng and Su, Dan and Xu, Yan and Ishii, Etsuko and Bang, Ye Jin and Madotto, Andrea and Fung, Pascale},
 journal = {ACM Computing Surveys},
 volume = {55},
 number = {12},
 articleno = {248},
 pages = {1--38},
 year = {2023},
 doi = {10.1145/3571730}
}

@misc{jiangSurveyLargeLanguage2024,
  title = {A {{Survey}} on {{Large Language Model Hallucination}} via a {{Creativity Perspective}}},
  author = {Jiang, Xuhui and Tian, Yuxing and Hua, Fengrui and Xu, Chengjin and Wang, Yuanzhuo and Guo, Jian},
  year = {2024},
  month = feb,
  number = {arXiv:2402.06647},
  eprint = {2402.06647},
  primaryclass = {cs},
  publisher = {arXiv},
  doi = {10.48550/arXiv.2402.06647},
  url = {http://arxiv.org/abs/2402.06647},
  urldate = {2025-08-11},
  archiveprefix = {arXiv}
}

@article{hao2024training,
  title={Training large language models to reason in a continuous latent space},
  author={Hao, Shibo and Sukhbaatar, Sainbayar and Su, DiJia and Li, Xian and Hu, Zhiting and Weston, Jason and Tian, Yuandong},
  journal={arXiv preprint arXiv:2412.06769},
  year={2024}
}

@inproceedings{li2025500xcompressor,
  title={500xcompressor: Generalized prompt compression for large language models},
  author={Li, Zongqian and Su, Yixuan and Collier, Nigel},
  booktitle={Proceedings of the 63rd Annual Meeting of the Association for Computational Linguistics (Volume 1: Long Papers)},
  pages={25081--25091},
  year={2025}
}

@article{ge2023context,
  title={In-context autoencoder for context compression in a large language model},
  author={Ge, Tao and Hu, Jing and Wang, Lei and Wang, Xun and Chen, Si-Qing and Wei, Furu},
  journal={arXiv preprint arXiv:2307.06945},
  year={2023}
}

@article{xia2025tokenskip,
  title={Tokenskip: Controllable chain-of-thought compression in llms},
  author={Xia, Heming and Leong, Chak Tou and Wang, Wenjie and Li, Yongqi and Li, Wenjie},
  journal={arXiv preprint arXiv:2502.12067},
  year={2025}
}

@article{zhou2025browsecomp,
  title={Browsecomp-zh: Benchmarking web browsing ability of large language models in chinese},
  author={Zhou, Peilin and Leon, Bruce and Ying, Xiang and Zhang, Can and Shao, Yifan and Ye, Qichen and Chong, Dading and Jin, Zhiling and Xie, Chenxuan and Cao, Meng and others},
  journal={arXiv preprint arXiv:2504.19314},
  year={2025}
}

@article{cheng2025glyph,
  title={Glyph: Scaling context windows via visual-text compression},
  author={Cheng, Jiale and Liu, Yusen and Zhang, Xinyu and Fei, Yulin and Hong, Wenyi and Lyu, Ruiliang and Wang, Weihan and Su, Zhe and Gu, Xiaotao and Liu, Xiao and others},
  journal={arXiv preprint arXiv:2510.17800},
  year={2025}
}

@article{phan2025humanity,
  title={Humanity's last exam},
  author={Phan, Long and Gatti, Alice and Han, Ziwen and Li, Nathaniel and Hu, Josephina and Zhang, Hugh and Zhang, Chen Bo Calvin and Shaaban, Mohamed and Ling, John and Shi, Sean and others},
  journal={arXiv preprint arXiv:2501.14249},
  year={2025}
}

@article{zhangHallucinationMitigationRetrievalAugmented2025,
 title = {Hallucination Mitigation for Retrieval-Augmented Large Language Models: A Review},
 author = {Zhang, Wan and Zhang, Jing},
 journal = {Mathematics},
 volume = {13},
 number = {5},
 articleno = {856},
 year = {2025},
 doi = {10.3390/math13050856}
}

@inproceedings{fanSurveyRAGMeeting2024,
 title = {A Survey on RAG Meeting LLMs: Towards Retrieval-Augmented Large Language Models},
 author = {Fan, Wenqi and Ding, Yujuan and Ning, Liangbo and Wang, Shijie and Li, Hengyun and Yin, Dawei and Chua, Tat-Seng and Li, Qing},
 booktitle = {Proceedings of the 30th ACM SIGKDD Conference on Knowledge Discovery and Data Mining (KDD)},
 pages = {6491--6501},
 year = {2024},
 doi = {10.1145/3637528.3671470}
}

@inproceedings{jiang2023llmlingua,
  title={LLMLingua: Compressing Prompts for Accelerated Inference of Large Language Models},
  author={Jiang, Huiqiang and Wu, Qianhui and Lin, Chin-Yew and Yang, Yuqing and Qiu, Lili},
  booktitle={The 2023 Conference on Empirical Methods in Natural Language Processing},
  year={2023}
}

@inproceedings{xu2024concise,
  title={Concise and Precise Context Compression for Tool-Using Language Models},
  author={Xu, Yang and Feng, Yunlong and Mu, Honglin and Hou, Yutai and Li, Yitong and Wang, Xinghao and Zhong, Wanjun and Li, Zhongyang and Tu, Dandan and Zhu, Qingfu and others},
  booktitle={Findings of the Association for Computational Linguistics ACL 2024},
  pages={16430--16441},
  year={2024}
}

@article{mu2024learning,
  title={Learning to compress prompts with gist tokens},
  author={Mu, Jesse and Li, Xiang and Goodman, Noah},
  journal={Advances in Neural Information Processing Systems},
  volume={36},
  year={2023}
}

@inproceedings{gecontext,
  title={In-context Autoencoder for Context Compression in a Large Language Model},
  author={Ge, Tao and Jing, Hu and Wang, Lei and Wang, Xun and Chen, Si-Qing and Wei, Furu},
  booktitle={The Twelfth International Conference on Learning Representations},
  year={2023}
}

@article{lei2025rhinoinsight,
  title={RhinoInsight: Improving Deep Research through Control Mechanisms for Model Behavior and Context},
  author={Lei, Yu and Si, Shuzheng and Wang, Wei and Wu, Yifei and Chen, Gang and Qi, Fanchao and Sun, Maosong},
  journal={arXiv preprint arXiv:2511.18743},
  year={2025}
}

@article{liu2025prosocial,
  title={Prosocial behavior in Large Language Models: Value alignment and affective mechanisms},
  author={Liu, Hao and Lei, Yu and Wu, Zhen},
  journal={Science China Technological Sciences},
  volume={68},
  number={8},
  pages={1820403},
  year={2025},
  publisher={Springer}
}

@article{liu2025context,
  title={Context cascade compression: Exploring the upper limits of text compression},
  author={Liu, Fanfan and Qiu, Haibo},
  journal={arXiv preprint arXiv:2511.15244},
  year={2025}
}

@misc{qwen3technicalreport,
      title={Qwen3 Technical Report}, 
      author={Qwen Team},
      year={2025},
      eprint={2505.09388},
      archivePrefix={arXiv},
      primaryClass={cs.CL},
      url={https://arxiv.org/abs/2505.09388}, 
}
\bibliographystyle{colm2024_conference}

\clearpage
\appendix
\section{Method Details}
\label{sec:appendix_method}

\begin{figure*}[h]
    \centering
    \includegraphics[width=\textwidth]{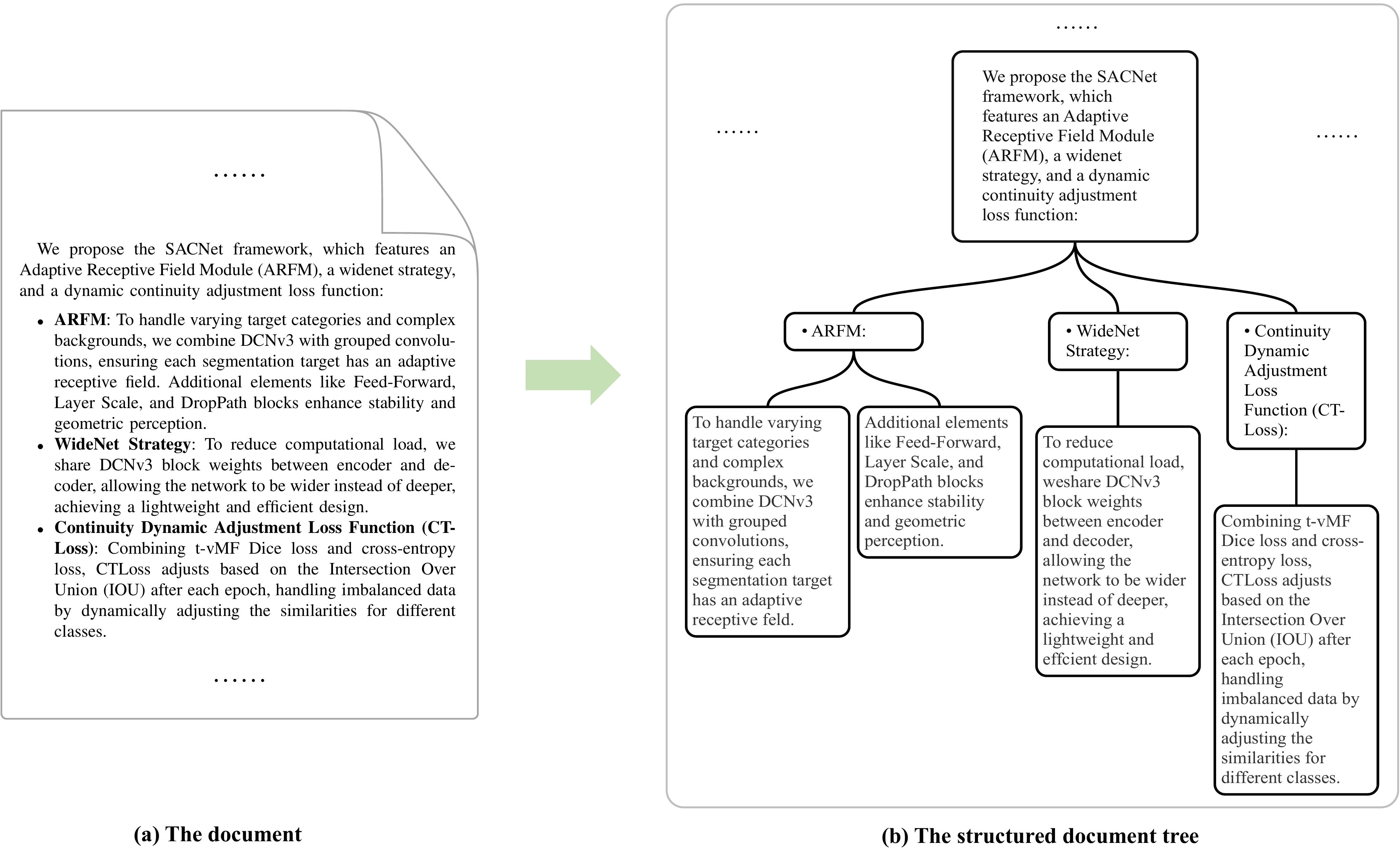}
    \caption{\textbf{Qualitative example of Traceable Context Compression.} 
    \textbf{(a) Input Document.} An example of a raw input document.
    \textbf{(b) The Structured Document Tree.} The model outputs a hierarchical index that reorganizes the linear EDUs into semantic clusters (Sections and Subsections). Note that the leaf nodes contain explicit coordinate pointers, allowing the system to faithfully retrieve the exact original text without hallucination.}
    \label{fig:details}
\end{figure*}

To provide a concrete understanding of our pipeline, we visualize the transformation process from linear text to a structured hierarchy in Figure \ref{fig:details}.

\section{Dataset Details}
\label{sec:dataset_details}

\subsection{StructBench}
Our \textbf{StructBench} dataset is constructed to ensure diversity in document structure, terminology, and formatting. The documents are sourced from the following 10 domains:
\begin{itemize}
\setlength\itemsep{0em} 
\item \textbf{Academic Papers/Journals:} Research articles with strict hierarchical structures.
\item \textbf{Government Documents:} Official files, policy mandates, and public announcements.
\item \textbf{Institutional/Business Reports:} Industry analysis, financial reports, and white papers.
\item \textbf{Technical Blogs:} Deep-dive articles on software, engineering, or science.
\item \textbf{Tutorials \& Guides:} Step-by-step instructional content.
\item \textbf{News:} Current events and journalistic reports.
\item \textbf{Opinion/Analysis:} Editorials, commentaries, and reviews.
\item \textbf{Popular Science:} Educational articles aimed at general audiences.
\item \textbf{Books:} Excerpts from non-fiction or structured literature.
\item \textbf{Lifestyle \& Entertainment:} Soft content including travel, hobbies, and arts.
\end{itemize}

\subsection{General Long-Context Understanding Domain}
To evaluate the generalization capabilities of our model beyond structural extraction, we conduct experiments on a diverse suite of general long-context tasks from the LongBench benchmark. This evaluation encompasses three distinct categories: Multi-Document QA, Summarization, and Few-shot Learning. The selected datasets cover a wide range of sources—including Wikipedia, government reports, and meeting transcripts—and support both English and Chinese languages.
As detailed in Table~\ref{tb:stat}, the context lengths vary significantly, ranging from approximately 2k to over 22k tokens, providing a rigorous testbed for assessing robustness in processing extensive unstructured contexts.

\begin{table*}[t]
\centering  
\resizebox{0.8\textwidth}{!}{
\begin{tabular}{llrccc}
\toprule
\textbf{Dataset} & \textbf{Source} & \textbf{Avg len} & \textbf{Metric} & \textbf{Language} & \textbf{\#data} \\
\midrule
\multicolumn{6}{l}{\emph{Multi-Document QA}} \\ 
HotpotQA & Wikipedia & 9,151 & F1 & English & 200 \\
2WikiMultihopQA & Wikipedia & 4,887 & F1 & English & 200 \\
MuSiQue & Wikipedia & 11,214 & F1 & English & 200 \\
DuReader & Baidu Search & 15,768 & Rouge-L & Chinese & 200 \\
\midrule
\multicolumn{6}{l}{\emph{Summarization}} \\
GovReport & Government report & 8,734 & Rouge-L & English & 200 \\
QMSum & Meeting & 10,614 & Rouge-L & English & 200 \\
MultiNews & News & 2,113 & Rouge-L & English & 200 \\
VCSUM & Meeting & 15,380 & Rouge-L & Chinese & 200 \\
\midrule
\multicolumn{6}{l}{\emph{Few-shot Learning}} \\
TREC & Web question & 5,177 & Accuracy (CLS) & English & 200 \\
TriviaQA & Wikipedia, Web & 8,209 & F1 & English & 200 \\
SAMSum & Dialogue & 6,258 & Rouge-L & English & 200 \\
LSHT & News & 22,337 & Accuracy (CLS) & Chinese & 200 \\
\bottomrule
\end{tabular}
}
\caption{An overview of the dataset statistics in LongBench used for evaluation. `Source' denotes the origin of the context. `Avg len' (average length) is computed using the number of words for English datasets and characters for Chinese datasets. `Accuracy (CLS)' refers to classification accuracy.}
\label{tb:stat}
\end{table*}

\subsection{DeepSearch Domain}

\textbf{High-Difficulty Reasoning (HLE):} We utilize \textit{Humanity’s Last Exam} (HLE)~\citep{phan2025humanity}, an expert-curated benchmark designed to assess frontier-level academic competence. From the original set of 2,500 highly challenging questions spanning multiple disciplines, we focus on the subset of 2,154 text-only questions to evaluate deep reasoning capabilities.

\noindent
\textbf{Real-World Noise (BrowseComp-ZH):} To test robustness in a messy information environment, we employ \textit{BrowseComp-ZH}~\citep{zhou2025browsecomp}. This is the first high-difficulty benchmark evaluating real-world web browsing and reasoning within the Chinese information ecosystem. It comprises 289 complex multi-hop queries across 11 domains (e.g., Film \& TV, Technology, Medicine) often embedded in noisy web layouts.

\section{Experiment  Details}
\label{sec:Experiment_appendix}
\subsection{Train Details}

\label{sec:train_model_details}
\begin{table}[h!] 
\centering
\small
\begin{tabular}{lr}
\toprule
\textbf{Hyperparameter} & \textbf{Value} \\
\midrule
\multicolumn{2}{l}{\textit{Model Configuration}} \\
Base Model & Qwen3-4B \\
Max Sequence Length & 32,768 \\
RoPE Base & 1,000,000 \\
Precision & bf16 \\
\midrule
\multicolumn{2}{l}{\textit{Optimization}} \\
Optimizer & AdamW \\
Optimizer Params & $\beta_1=0.9, \beta_2=0.95$ \\
Peak Learning Rate & $1 \times 10^{-5}$ \\
Min Learning Rate & $1 \times 10^{-6}$ \\
LR Scheduler & Cosine \\
Warmup Ratio & 0.1 \\
Weight Decay & 0.1 \\
Gradient Clipping & 0.5 \\
\midrule
\multicolumn{2}{l}{\textit{Batching \& Parallelism}} \\
Global Batch Size & 128 \\
Training Iterations & 1,296 \\
Tensor Parallelism (TP) & 4 \\
Sequence Parallelism & True \\
\bottomrule
\end{tabular}
\caption{Hyperparameters and configuration used for training the LingEDU.}
\label{tab:train_hyperparams}
\end{table}

We present the detailed training configuration in Table~\ref{tab:train_hyperparams}. The model is fine-tuned based on the Qwen3-4B architecture. To accommodate the long-context requirement, we extend the maximum sequence length to 32,768 tokens and adjust the Rotary Positional Embedding (RoPE) base frequency to $1,000,000$. We utilize the Adam optimizer with $\beta_1=0.9$, $\beta_2=0.95$, and $\epsilon=1\text{e-}8$. A weight decay of $0.1$ and a gradient clipping threshold of $1.0$ are applied to stabilize training. The learning rate follows a cosine decay schedule, starting from a peak of $1\text{e-}5$ and decaying to a minimum of $1\text{e-}6$, with a linear warmup phase covering the first 10\% of the training steps.

\subsection{Detailed Model Specifications}
\label{sec:appendix_model_details}

To ensure the reproducibility of our experiments, Table \ref{tab:appendix_model_versions} lists the specific version identifiers and access paths for all models used in the \textit{Structure Extraction} and \textit{Downstream Long-Context} tasks.

\begin{table*}[h]
    \centering
    \small
    \renewcommand{\arraystretch}{1.3}
    \resizebox{0.8\textwidth}{!}{
    \begin{tabular}{l|c|l} 
    \toprule
    \textbf{Model Name in Paper} & \textbf{Experiment Scope} & \textbf{Real API ID / Checkpoint / Address} \\
    \midrule
    \multicolumn{3}{l}{\cellcolor{gray!10}\textit{\textbf{OpenAI Models}}} \\
    GPT-4o & Structure & \texttt{gpt-4o-2024--11-20} \\
    GPT-4.1 & Structure & \texttt{gpt-4.1-2025-04-14} \\
    OpenAI o3 & Structure & \texttt{o3-2025-04-16} \\
    OpenAI o4-mini & Structure & \texttt{o4-mini-2025-04-16} \\
    GPT-5 & \textbf{Downstream} & \texttt{gpt-5-2025-08-07} \\
    \midrule
    \multicolumn{3}{l}{\cellcolor{gray!10}\textit{\textbf{Anthropic Models}}} \\
    Claude-3.7-Sonnet & Structure & \texttt{claude-3-7-sonnet-20250219} \\
    Claude-4 & Structure & \texttt{claude-sonnet-4-20250514} \\
    Claude Opus 4.1 & \textbf{Downstream} & \texttt{claude-opus-4-1-20250805} \\
    \midrule
    \multicolumn{3}{l}{\cellcolor{gray!10}\textit{\textbf{Google Models}}} \\
    Gemini-2.5-flash & Structure & \texttt{gemini-2.5-flash} \\
    Gemini-2.5-pro & Structure & \texttt{gemini-2.5-pro} \\
    Gemini 3 Pro & \textbf{Downstream} & \texttt{gemini-3-pro-preview} \\
    \midrule
    \multicolumn{3}{l}{\cellcolor{gray!10}\textit{\textbf{DeepSeek Models}}} \\
    DeepSeek-V3 & Structure & \texttt{deepseek-v3-250324} \\
    DeepSeek-R1 & Structure, \textbf{Downstream} & \texttt{deepseek-r1-250528} \\
    DeepSeek-V3.1 & \textbf{Downstream} & \texttt{deepseek-v3-1-250821} \\
    DeepSeek-V3.2 & \textbf{Downstream} & \texttt{deepseek-v3-2-251201} \\
    \midrule
    \multicolumn{3}{l}{\cellcolor{gray!10}\textit{\textbf{Qwen Models (Local / Open Weights)}}} \\
    Qwen3-32B & Structure & \texttt{https://huggingface.co/Qwen/Qwen3-32B} \\
    Qwen3-235B & Structure, \textbf{Downstream} & \texttt{https://huggingface.co/Qwen/Qwen3-235B-A22B} \\
    \midrule
    \multicolumn{3}{l}{\cellcolor{gray!10}\textit{\textbf{Specialized Tools}}} \\
    Jina-Reader & Structure & \texttt{https://jina.ai/reader} \\
    Firecrawl & Structure & \texttt{https://www.firecrawl.dev} \\
    \bottomrule
    \end{tabular}
    }
    \caption{Detailed version tracking for all models. ``Structure'' denotes models used in Table 1 (StructBench), while ``Downstream'' refers to the Long-Context evaluation tasks. Specific identifiers or URLs are provided in the third column to specify the exact model artifacts used.}
    \label{tab:appendix_model_versions}
\end{table*}

\subsection{Robustness Analysis Across Context Lengths}
\label{sec:length_exp1}

\begin{figure*}[t]
    \centering
    \includegraphics[width=0.8\textwidth]{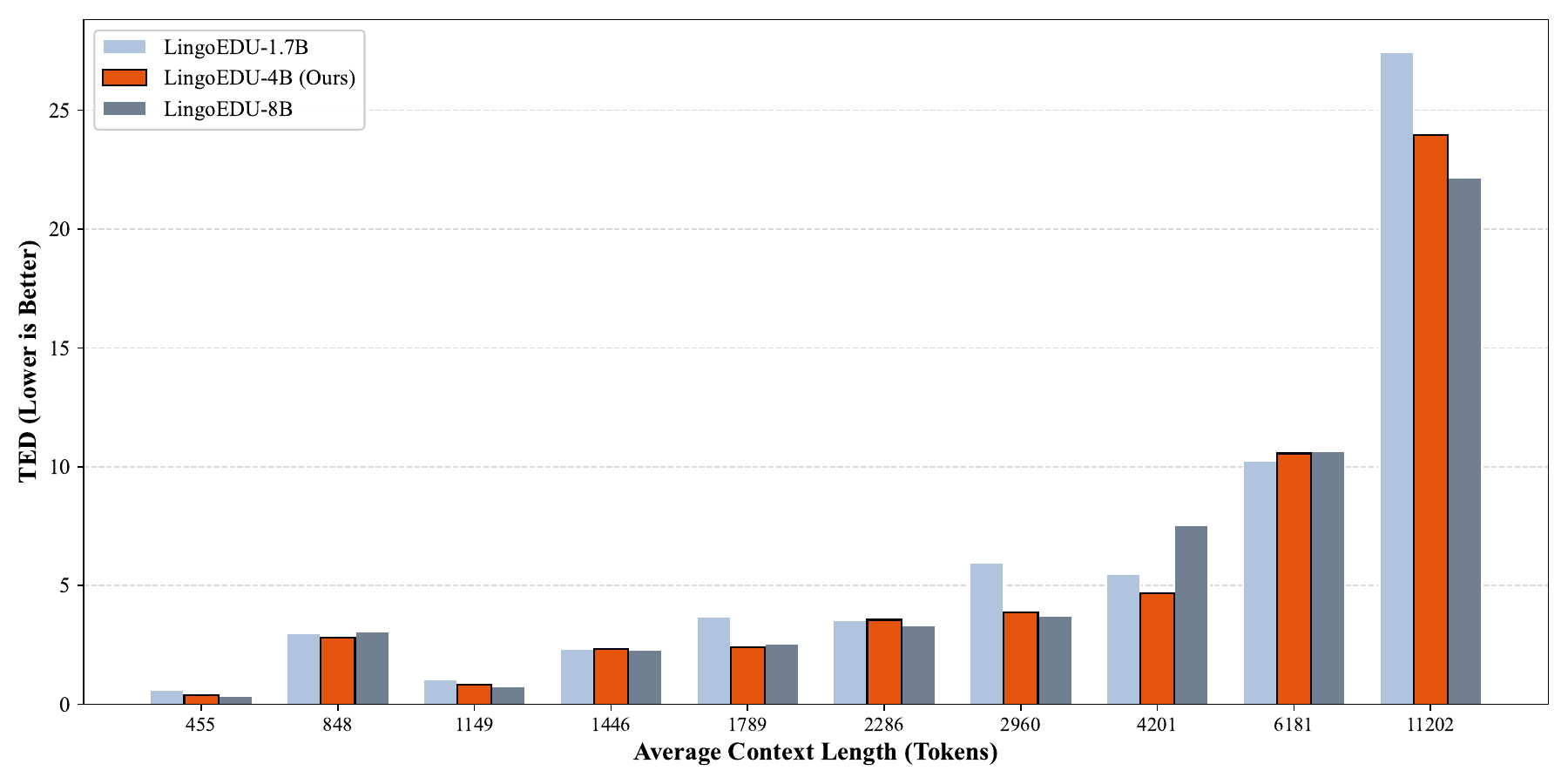}
    \caption{Performance comparison (TED) on StructBench across varying context lengths. The dataset is divided into 10 intervals by token count. Lower TED indicates better performance. The LingoEDU-4B model (orange) demonstrates superior efficiency-robustness balance. It consistently outperforms the 1.7B model in long contexts and achieves comparable—or even superior (e.g., Bin 7)—structural consistency relative to the larger 8B model, validating its selection as the primary backbone.}
    \label{fig:length_analysis}
\end{figure*}

To evaluate the model's capability in maintaining structural constraints over extended inputs, we partition the 248 StructBench documents into 10 bins based on token length. We employ Tree Edit Distance (TED) as the primary metric, where lower scores indicate higher structural fidelity. 

As depicted in Figure~\ref{fig:length_analysis}, all models maintain low error rates in short contexts (Avg $<3$k tokens). However, distinct performance characteristics emerge as the context length extends. The LingoEDU-4B model demonstrates remarkable resilience, effectively bridging the gap between lightweight and large-scale architectures. 

Notably, in the medium-to-long range (e.g., Bin 7, Avg $\sim$4.2k tokens), LingoEDU-4B achieves a TED of 4.68, outperforming both the 1.7B baseline and remarkably the 8B variant (TED 7.52). Furthermore, in the most challenging regime (Bin 9, Avg $>$11k tokens), while the 1.7B model suffers significant degradation (TED 27.44), LingoEDU-4B maintains a competitive performance (TED 23.96), rivaling the stability of the 8B model. This indicates that LingoEDU-4B offers the optimal trade-off, delivering 8B-level long-context robustness with significantly higher inference efficiency.

\subsection{Cost Analysis}
\label{sec:appendix_cost_analysis}

To evaluate the economic efficiency of our proposed method, we conducted a comprehensive cost comparison between the pure LLM-based pipeline (Baseline) and our LingoEDU-integrated pipeline.

\paragraph{Pricing Model Assumptions.}
The cost calculations are based on the pricing of \textit{GPT-4.1} . The pricing scheme is \textbf{\$2.00 per 1M input tokens} and \textbf{\$8.00 per 1M output tokens}. 
For the baseline, the LLM handles the entire specific parsing and answering workflow.
For our method, the parsing is offloaded to the local LingoEDU-4B model. While local inference is not free due to hardware amortization and electricity, it is significantly cheaper. Based on our deployment statistics, the cost for processing a single document with LingoEDU is approximately \textbf{\$0.0007}, compared to \textbf{\$0.0168} with GPT-4.1. This $\sim$24x cost efficiency allows LingoEDU to process massive amounts of tokens locally with minimal economic impact.

\paragraph{Cost Breakdown.}
Table~\ref{tab:cost_comparison} details the token consumption and estimated expenses.


\begin{table*}[h]
\centering
\small
\renewcommand{\arraystretch}{1.2}
\setlength{\tabcolsep}{4pt}

\begin{tabularx}{0.8\textwidth}{@{} l l >{\raggedleft\arraybackslash}X >{\raggedleft\arraybackslash}X >{\raggedleft\arraybackslash}X @{}}
\toprule
\textbf{Stage} & \textbf{Metric} & \textbf{Direct LLM} & \textbf{LLM Pipeline} & \textbf{Ours (LingoEDU)} \\ 
\midrule
\multicolumn{5}{l}{\textit{\textbf{1. Parsing Phase}}} \\
& Method & - & GPT-4.1 Gen. & LingoEDU (Local) \\
& Input Tokens & - & 5,955,972 & 5,955,972 \\
& Output Tokens & - & 1,314,406 & 2,170,766 \\
& \textbf{Est. Cost} & \textbf{-} & \textbf{\$22.43} & \textbf{\$0.53} \\
\midrule
\multicolumn{5}{l}{\textit{\textbf{2. Reranking Phase}}} \\
& Method & - & Qwen3-0.6B & Qwen3-0.6B \\
& Tokens & - & 1,013,704 & 2,170,766 \\
& \textbf{Est. Cost} & \textbf{-} & \textbf{<\$0.01} & \textbf{<\$0.01} \\
\midrule
\multicolumn{5}{l}{\textit{\textbf{3. Answering Phase}}} \\
& Method & GPT-4.1 & GPT-4.1 & GPT-4.1 \\
& Input Tokens & 5,955,972 & 147,995 & 2,605,437 \\
& Output Tokens & 1,357 & 1,157 & 1,475 \\
& \textbf{Est. Cost} & \textbf{\$11.92} & \textbf{\$0.31} & \textbf{\$5.22} \\
\midrule
\textbf{Total} & \textbf{Total Cost} & \textbf{\$11.92} & \textbf{\$22.74} & \textbf{\$5.76} \\
& \textbf{Cost Comparison} & \textbf{+107\%} & \textbf{+295\%} & \textbf{Base} \\
\bottomrule

\multicolumn{5}{p{0.78\textwidth}}{\footnotesize \textbf{Pricing}: GPT-4.1 (\$2.00/1M Input, \$8.00/1M Output). Local Reranker (Qwen3-0.6B) cost is negligible (<\$0.002).} \\
\multicolumn{5}{p{0.78\textwidth}}{\footnotesize \textbf{1}: Direct LLM processes raw tokens directly. \textbf{2}: LLM Pipeline is expensive due to generating 1.3M tokens during parsing.} \\
\end{tabularx}
\caption{Cost comparison across three strategies. We use a strictly constrained layout. \textbf{Direct LLM} incurs high input costs. The \textbf{LLM Pipeline} is the most expensive due to generation costs. \textbf{Ours (LingoEDU)} achieves the lowest cost.}
\label{tab:cost_comparison}
\end{table*}

\paragraph{Analysis.}
As shown in Table~\ref{tab:cost_comparison}, our strategy achieves the lowest total cost (\$5.76), representing a \textbf{51.7\% reduction} compared to the Direct LLM approach (\$11.92) and a massive \textbf{74.7\% reduction} compared to the LLM-based Pipeline (\$22.74). The data highlights two critical economic advantages:

\begin{enumerate}
    \item \textbf{Avoiding the "Generation Tax" in Parsing:} The LLM Pipeline incurs prohibitively high costs during the Parsing Phase (\$22.43) because it relies on GPT-4.1 to generate structured outputs. Generating 1.3M output tokens triggers the expensive prediction rate (\$8/M). In contrast, \textbf{Ours (LingoEDU)} offloads this heavy structural extraction to local models. This allows us to process the same 5.9M raw tokens for virtually zero cost (\$0.53 overhead), completely bypassing commercial API fees for the most data-intensive stage.
    
    \item \textbf{Strategic Context Allocation:} 
    Compared to "Direct LLM," which blindly feeds all 5.9M raw tokens into the costly API (\$11.92), our method uses the local Qwen3 reranker to filter noise efficiently. This reduces the final input volume to 2.6M tokens, cutting input costs by half while maintaining high information density. 
    Conversely, compared to the "LLM Pipeline" (which aggressively reduces context to 148k tokens to save money), we reinvest the savings from the parsing phase into a much richer context (2.6M tokens) for the Answer Phase. This strikes an optimal balance: providing significantly more context than the baseline pipeline to ensure accuracy, while remaining far cheaper than direct processing.
\end{enumerate}

\subsection{Impact of Retrieval Granularity on Compression and Performance}
\label{sec:com_performance}

To determine the optimal granularity for our structure-aware retrieval, we conduct an ablation study on LongBench by varying the number of retrieved nodes $k$ (e.g., $k \in \{3, 10, 20\}$). We analyze the trade-off between task performance (Metric Score) and computational efficiency (Compression Rate), with the standard Top-100 retrieval serving as the baseline.

\begin{figure*}[t]
    \centering
    \includegraphics[width=\textwidth]{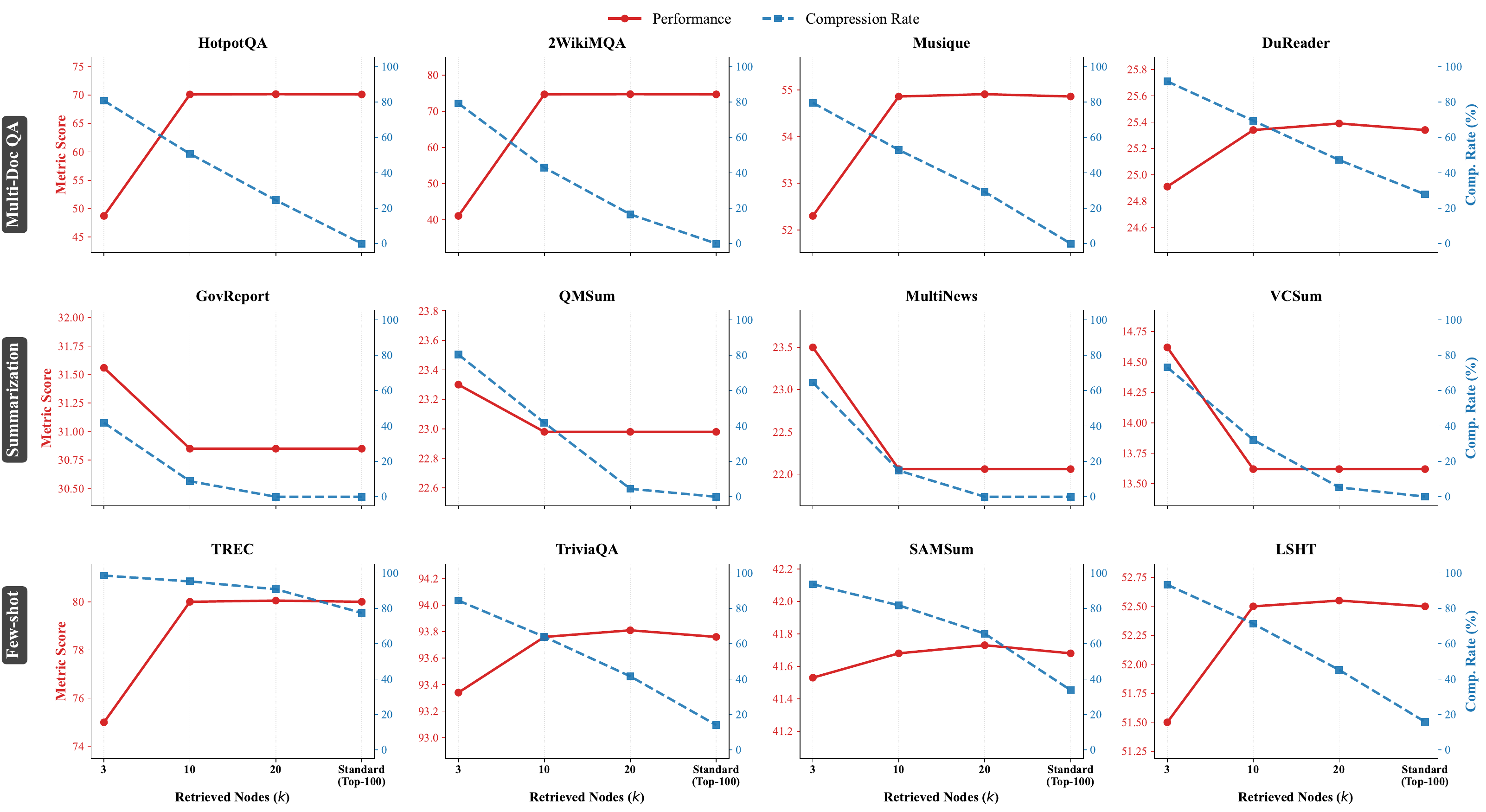}
    \caption{Ablation analysis of retrieval size ($k$) versus performance and compression on LongBench tasks. The left y-axis denotes the metric score, while the right y-axis shows the input compression rate. The dashed line represents the Standard (Top-100) baseline. We observe that $k=10$ represents the optimal trade-off: it achieves performance nearly identical to or better than the dense baseline while maintaining a significantly higher compression rate. Increasing $k$ further to 20 provides negligible metric improvements but incurs a steeper cost in context length.
}
    \label{fig:ablation_topk}
\end{figure*}

As illustrated in Figure~\ref{fig:ablation_topk}, increasing $k$ naturally improves performance by incorporating more context; however, it simultaneously reduces the compression rate, leading to higher computational overhead.

Justification for Choosing $k=10$. Our empirical results identify $k=10$ as the optimal operating point.
\begin{itemize}
    \item \textbf{High Fidelity:} At $k=10$, the model performs comparably to, and in some datasets (e.g., HotpotQA, Musique) even surpasses, the Standard baseline. This indicates that the top-10 identified nodes capture the vast majority of the task-relevant signal effectively suppressing noise found in the larger top-100 context.
    \item \textbf{Efficiency Gain:} While further increasing $k$ to 20 yields only marginal performance gains (diminishing returns), $k=10$ maintains a significantly higher compression rate (preserving $>85\%$ compression on average). 
\end{itemize}
Consequently, we adopt $k=10$ as the default setting for our method, as it strikes the most favorable balance between maximizing structural accuracy and minimizing token consumption.

\section{Prompt Template}
\label{sec:prompt_all}

\subsection{LLM Baselines on StructBench}
\label{sec:appendix_prompt}

To strictly pinpoint the structural understanding capabilities of general-purpose Large Language Models (LLMs), we utilized a unified zero-shot system prompt. This prompt essentially instructs the model to act as a parser, extracting the hierarchy without modifying the content.

The specific prompt content is visually presented in Figure~\ref{fig:prompt_llm_baselines}.

It is important to clarify the distinct inference paradigms used in our experiments:
\begin{itemize}
    \item \textbf{LLM-based Baselines (Applied):} This prompt was applied to all general-purpose models listed in Table~\ref{sec:appendix_model_details}, including GPT-4o series, Claude series, DeepSeek series, and Qwen series.
    \item \textbf{Commercial APIs (Not Applied):} Services like \textit{Jina Reader} and \textit{Firecrawl} operate as specialized black-box parsers. They ingest URLs or files and return structure via internal logic, rendering external prompting inapplicable.
    \item \textbf{EDU (Ours) (Not Applied):} Unlike general-purpose LLMs that require detailed instructions (Prompt Engineering) to define the task, our model is explicitly trained via Supervised Fine-Tuning (SFT) for this specific objective. It accepts the raw document stream and outputs the structured tree end-to-end, relying on its internal parametric knowledge rather than prompt-based context.
\end{itemize}

\begin{figure*}[h]
\centering
\begin{tcolorbox}[title=System Prompt used for LLM Baselines on StructBench, width=\textwidth, colback=gray!5, colframe=black, fonttitle=\bfseries]
\small
\textbf{Instruction:} \\
Work your way down through the article's heading structure, outputting each level of heading in Markdown format. \\

- First, use only the original headings—do not include body text, do not summarize, and do not rewrite. \\

- If a heading is split into multiple sentences due to punctuation, combine them into a single complete heading and output it as one entry. \\

- Output the heading structure top-down, carefully identifying hierarchical relationships and determining whether overly detailed levels are necessary to output. \\

- Finally, return only the final result without any additional explanations. \\

Article: [INPUT CONTENT]

\end{tcolorbox}
\caption{Unified system prompt used for zero-shot baseline evaluation.}
\label{fig:prompt_llm_baselines}
\end{figure*}

\subsection{LLM Baselines on LongBench}

To assess the effectiveness of our proposed framework on LongBench, specifically for the entry labeled \textbf{Ours (LingoEDU)} in Table~\ref{tab:main_results}, we implemented a two-stage retrieval-augmented generation pipeline. This pipeline leverages LingoEDU for structural parsing and a ranking model for context selection.
The specific prompts for both stages are visually presented in Figure~\ref{fig:prompt_lingoedu}.
It is crucial to understand how the components interact in our experiment:
\begin{itemize}
    \item \textbf{LingoEDU (Structure Parsing):} First, the raw long context is processed by LingoEDU, which segments the text into a hierarchical tree of Elementary Discourse Units (EDUs). This assigns a unique index ID and a depth level to every meaningful span of text.
    \item \textbf{Ranking Model (Context Selection):} For the QA stage, we do not feed the entire document to the LLM. Instead, a lightweight ranking model scores the relevance of each EDU against the user query. We select the top-$k$ relevant nodes to construct the \texttt{\{ctxt\}} variable, ensuring adherence to the token budget while maintaining high information density.
    \item \textbf{LLM (Reasoning \& Synthesis):} The general-purpose LLM acts as the final reasoner. Crucially, it is instructed to cite the specific node indices (e.g., \texttt{[12]}) provided by LingoEDU, allowing for traceable answers grounded in the retrieved segments.
\end{itemize}
\begin{figure*}[h]
\centering
\begin{tcolorbox}[title=Stage 1: Content Summarization Prompt (Indexing Phase), width=\textwidth, colback=gray!5, colframe=black, fonttitle=\bfseries]
\small
\textbf{System Message:} \\
You are a professional content analyst. Please always output valid JSON.
\textbf{User Prompt:} \\
Please generate a professional retrieval content based on the following:
\begin{itemize}
    \item Source: \texttt{\{source\_desc\}}
    \item Title: \texttt{\{title\}} (or 'No explicit title detected')
    \item Hierarchical Content: \\
    \texttt{\{content\_text\}} \textit{(Note: Formatted with indentation strings matching EDU levels)}
\end{itemize}
\textbf{Summarization Requirements:} \\
1) Provide a 150-250 word summary. \\
2) List 3-5 key points. \\
3) Outline the main purpose/function. \\
4) Briefly describe content structure characteristics.
\textbf{Output JSON Format:} \\
\texttt{\{ "summary": "...", "key\_points": ["..."], "main\_purpose": "...", "content\_structure": "...", "information\_value": "High/Medium/Low" \}}
\end{tcolorbox}
\vspace{0.3cm}
\begin{tcolorbox}[title=Stage 2: Retrieval-Augmented QA Prompt (Inference Phase), width=\textwidth, colback=gray!5, colframe=black, fonttitle=\bfseries]
\small
\textbf{System Message:} \\
You are a rigorous retrieval QA assistant.
\textbf{User Prompt:} \\
You are a rigorous retrieval QA assistant. Answer only based on the provided context. Do not fabricate information.
\textbf{Question:} \\
\texttt{\{query\}}
\textbf{Context (indexed by node ID):} \\
\texttt{\{ctxt\}} \\
\textit{(Format: `[0] Text... [5] Text...` — selected by the Ranking Model)}
\textbf{Please provide:}
\begin{itemize}
    \item Direct answer (if derivable).
    \item Concise explanation (based on the context).
    \item Citations of the node indices used (e.g., \texttt{[12, 15]}).
\end{itemize}
\textbf{Requirements:} \\
- If the context is insufficient, explicitly state "Insufficient to answer". \\
- Do not introduce information outside the provided context.
\end{tcolorbox}
\caption{Unified prompts used for the \textbf{Ours (LingoEDU)} pipeline in LongBench. Stage 1 summarizes the structured EDU tree, while Stage 2 performs citation-aware QA using ranked EDU nodes.}
\label{fig:prompt_lingoedu}
\end{figure*}

\subsection{LLM Baselines on DeepSearch}

\begin{figure*}[h]
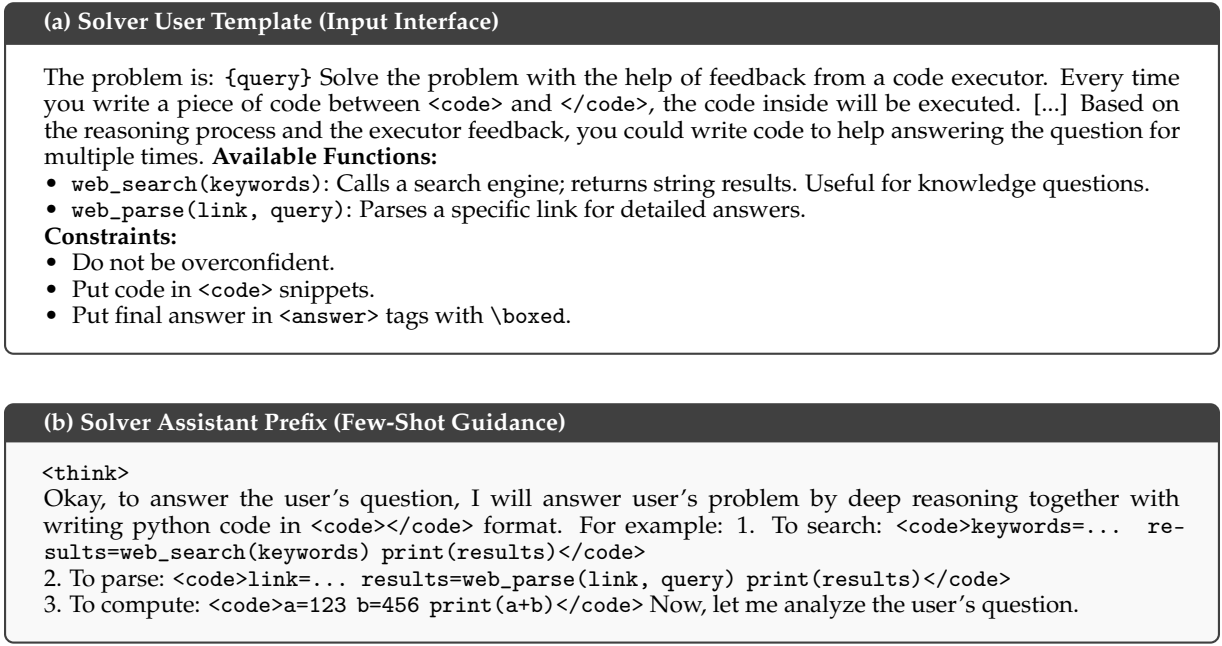

\centering
\begin{tcolorbox}[
    title={\small \textbf{(a) Solver User Template} (Input Interface)}, 
    width=\textwidth, 
    colback=white, 
    colframe=black!75, 
    fonttitle=\bfseries,
    boxrule=0.8pt
]
\footnotesize
The problem is: \texttt{\{query\}}
Solve the problem with the help of feedback from a code executor. Every time you write a piece of code between \texttt{<code>} and \texttt{</code>}, the code inside will be executed. [...] Based on the reasoning process and the executor feedback, you could write code to help answering the question for multiple times.
\textbf{Available Functions:}
\begin{itemize}[leftmargin=*, nosep]
    \item \texttt{web\_search(keywords)}: Calls a search engine; returns string results. Useful for knowledge questions.
    \item \texttt{web\_parse(link, query)}: Parses a specific link for detailed answers.
\end{itemize}
\textbf{Constraints:}
\begin{itemize}[leftmargin=*, nosep]
    \item Do not be overconfident.
    \item Put code in \texttt{<code>} snippets.
    \item Put final answer in \texttt{<answer>} tags with \texttt{\textbackslash boxed}.
\end{itemize}
\end{tcolorbox}
\vspace{0.15cm}
\begin{tcolorbox}[
    title={\small \textbf{(b) Solver Assistant Prefix} (Few-Shot Guidance)}, 
    width=\textwidth, 
    colback=gray!5, 
    colframe=black!75, 
    fonttitle=\bfseries,
    boxrule=0.8pt
]
\footnotesize
\texttt{<think>}\\
Okay, to answer the user's question, I will answer user's problem by deep reasoning together with writing python code in \texttt{<code></code>} format. For example:
1. To search: \texttt{<code>keywords=... results=web\_search(keywords) print(results)</code>} \\
2. To parse: \texttt{<code>link=... results=web\_parse(link, query) print(results)</code>} \\
3. To compute: \texttt{<code>a=123 b=456 print(a+b)</code>}
Now, let me analyze the user's question.
\end{tcolorbox}
\caption{Prompts used for the \textbf{Solver Agent}. The user template (a) defines the tool-use environment, while the assistant prefix (b) primes the model for Chain-of-Thought reasoning paired with Python code execution.}
\label{fig:solver_prompts}
\end{figure*}

\begin{figure*}[h]
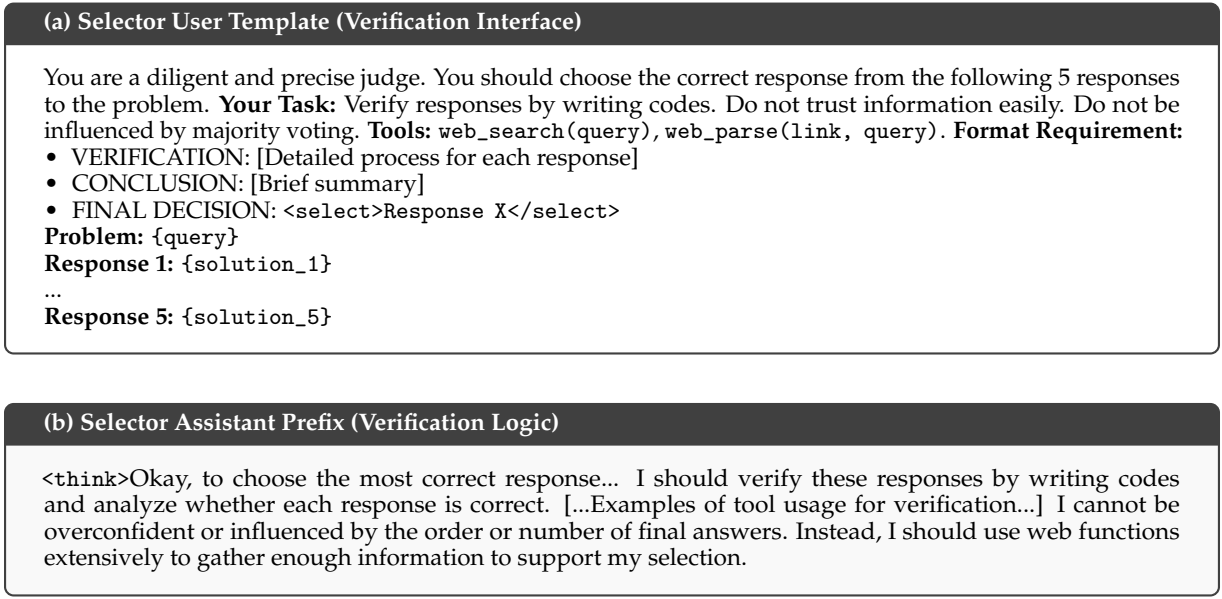

\centering
\begin{tcolorbox}[
    title={\small \textbf{(a) Selector User Template} (Verification Interface)}, 
    width=\textwidth, 
    colback=white, 
    colframe=black!75, 
    fonttitle=\bfseries,
    boxrule=0.8pt
]
\footnotesize
You are a diligent and precise judge. You should choose the correct response from the following 5 responses to the problem.
\textbf{Your Task:} Verify responses by writing codes. Do not trust information easily. Do not be influenced by majority voting.
\textbf{Tools:} \texttt{web\_search(query)}, \texttt{web\_parse(link, query)}.
\textbf{Format Requirement:}
\begin{itemize}[leftmargin=*, nosep]
    \item VERIFICATION: [Detailed process for each response]
    \item CONCLUSION: [Brief summary]
    \item FINAL DECISION: \texttt{<select>Response X</select>}
\end{itemize}
\textbf{Problem:} \texttt{\{query\}} \\
\textbf{Response 1:} \texttt{\{solution\_1\}} \\
... \\
\textbf{Response 5:} \texttt{\{solution\_5\}}
\end{tcolorbox}
\vspace{0.15cm}
\begin{tcolorbox}[
    title={\small \textbf{(b) Selector Assistant Prefix} (Verification Logic)}, 
    width=\textwidth, 
    colback=gray!5, 
    colframe=black!75, 
    fonttitle=\bfseries,
    boxrule=0.8pt
]
\footnotesize
\texttt{<think>}Okay, to choose the most correct response... I should verify these responses by writing codes and analyze whether each response is correct.
[...Examples of tool usage for verification...]
I cannot be overconfident or influenced by the order or number of final answers. Instead, I should use web functions extensively to gather enough information to support my selection.
\end{tcolorbox}
\caption{Prompts used for the \textbf{Selector Agent}. This stage employs a "Judge" persona that critically evaluates five candidate solutions using independent tool calls before making a final selection.}
\label{fig:selector_prompts}
\end{figure*}

To handle complex queries requiring multi-step reasoning and verification, our DeepSearch framework employs a hierarchical prompting strategy. This strategy coordinates two distinct agent roles: the \textbf{Solver} (which generates candidate solutions using tools) and the \textbf{Selector} (which verifies and chooses the best solution). Additionally, we implement a \textbf{Search Enhancement} module that injects structured retrieval results into the reasoning process. 
The prompt designs for these components are detailed below.
\begin{itemize}
    \item \textbf{Figure \ref{fig:solver_prompts}:} Shows the prompts for the Solver agent, enabling code execution and web interactions.
    \item \textbf{Figure \ref{fig:selector_prompts}:} Shows the prompts for the Selector agent, enforcing strict verification protocols.
    \item \textbf{Figure \ref{fig:search_enhancement}:} Illustrates how raw search results are processed via EDU parsing and LLM summarization before injection.
\end{itemize}

\begin{figure*}[h]
\centering
\begin{tcolorbox}[
    title={\small \textbf{Intelligent Summary Generation based on EDU Parsing}}, 
    width=\textwidth, 
    colback=white, 
    colframe=black!75, 
    fonttitle=\bfseries,
    boxrule=0.8pt
]
\footnotesize
\textbf{System:} You are a professional search result analyst... always return valid JSON.
\textbf{User Prompt:} \\
Please generate a professional summary for the search result based on the hierarchical content structure:
\begin{itemize}[leftmargin=*, nosep]
    \item \textbf{Query:} \texttt{\{query\}}
    \item \textbf{URL / Title:} \texttt{\{url\}} / \texttt{\{title\}}
    \item \textbf{Hierarchical Content:} \texttt{\{main\_content\}} (from EDU parsing)
    \item \textbf{Extracted Key Points:} \texttt{\{key\_points\}}
\end{itemize}
\textbf{Requirements:}
1. Analyze relevance to query. 2. Concise summary (100-200 words). 3. Highlight relevant info. 4. Identify 3-5 key points. 5. Evaluate credibility.
\textbf{Output JSON:}
\texttt{
\{ "summary": "...", "key\_points": ["..."], "relevance\_score": "...", "content\_quality": "...", "main\_topics": ["..."] \}
}
\end{tcolorbox}
\caption{Search Enhancement Prompt. When EDU parsing is enabled, this prompt converts raw hierarchical web content into a structured, relevance-scored JSON summary, which is then injected into the Solver's context.}
\label{fig:search_enhancement}
\end{figure*}

\end{document}